\documentclass[10pt,journal,compsoc]{IEEEtran}
\ifCLASSOPTIONcompsoc
  \usepackage[nocompress]{cite}
\else
  \usepackage{cite}
\fi
\usepackage[pagebackref=true,breaklinks=true,letterpaper=true,bookmarks=false]{hyperref}

\usepackage{amsmath,amssymb,amsfonts}
\usepackage{algorithmic}
\usepackage{graphicx}
\usepackage{textcomp}
\usepackage{xcolor}
\usepackage{multirow}
\usepackage{arydshln}
\usepackage{subcaption}
\usepackage{slashbox}

\newcommand{\REVISED}[1]{{\color{black}{#1}}}
\newcommand{\topone}[1]{{\underline{\textbf{#1}}}}

\ifCLASSINFOpdf
\else
\fi
\begin{document}
%
\title{DeepTag: A General Framework for \\Fiducial Marker Design and Detection}
%
%
%
%

\author{Zhuming Zhang, 
        Yongtao Hu, 
        Guoxing Yu,
        Jingwen Dai 
  \thanks{Manuscript received 2021.}}
\markboth{IEEE TRANSACTIONS ON PATTERN ANALYSIS AND MACHINE INTELLIGENCE,~Vol.~X, No.~X, XX XXXX}%
{Shell \MakeLowercase{\textit{et al.}}: Bare Demo of IEEEtran.cls for Computer Society Journals}
%



\IEEEtitleabstractindextext{%

\begin{abstract}
    A fiducial marker system usually consists of markers, a detection algorithm, and a coding system. 
    The appearance of markers and the detection robustness are generally limited by the existing detection algorithms, which are hand-crafted with traditional low-level image processing techniques.
    Furthermore, a sophisticatedly designed coding system is required to overcome the shortcomings of both markers and detection algorithms.
    To improve the flexibility and robustness in various applications, we propose a general deep learning based framework, \textit{DeepTag}, for fiducial marker design and detection.
    DeepTag not only supports detection of a wide variety of existing marker families, but also makes it possible to design new marker families with customized local patterns.
    Moreover, we propose an effective procedure to synthesize training data on the fly without manual annotations.
    Thus, DeepTag can easily adapt to existing and newly-designed marker families.
    To validate DeepTag and existing methods, beside existing datasets, we further collect a new large and challenging dataset where markers are placed in different view distances and angles. 
    Experiments show that DeepTag well supports different marker families and greatly outperforms the existing methods in terms of both detection robustness and pose accuracy.
    Both code and dataset are available at \url{https://herohuyongtao.github.io/research/publications/deep-tag/}.
\end{abstract}

\begin{IEEEkeywords}
Fiducial Marker, Deep Learning, Object Detection, Marker Design, Monocular Pose Estimation
\end{IEEEkeywords}}

\maketitle

\IEEEdisplaynontitleabstractindextext

%
\IEEEpeerreviewmaketitle

\IEEEraisesectionheading{\section{Introduction}\label{sec:introduction}}
\IEEEPARstart{F}{iducial} markers are artificial landmarks that provide correspondence between 2D images and a known 3D model.
By adding them to a scene, pose estimation can be achieved in applications like augmented reality, user pose input with handheld devices, and robot navigation, etc.
Superior to markerless computer vision that directly analyzes natural objects, fiducial markers offer reliable features of sufficient quantity and uniqueness.

\begin{figure}[tp]
	\centering
	\includegraphics[width=0.49\textwidth]{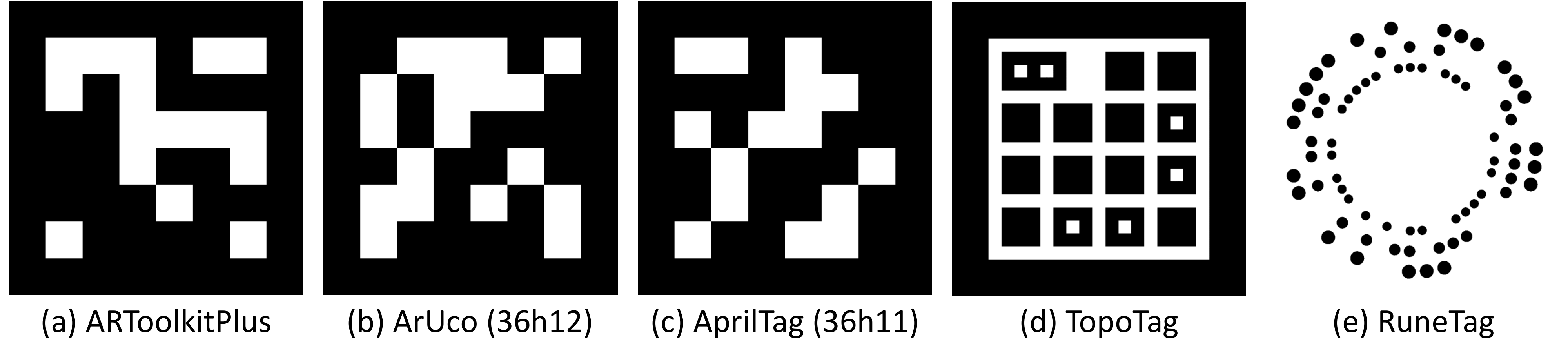}
	\vspace{-0.2in}
	\caption{Existing fiducial markers. (a)-(c) Common checkerboard markers. (d) Checkerboard marker with fixed regions. (e) Marker with dense dots arranged on circles.}
	\label{fig:example_markers}
	\vspace{-0.1in}
\end{figure}

To compute a unique pose from a single marker, at least four non-collinear \emph{keypoints} are required.
On the other hand, a \emph{library} that consists of different markers is required to distinguish multiple objects.
For instance, checkerboard markers in Fig.~\ref{fig:example_markers}(a-c) have a quadrilateral boundary that provides four corner points for pose estimation.
Black-and-white cells arranged in a regular grid represent \emph{digital symbols} "0" or "1", and predefined libraries map bit sequences to marker IDs.

A fiducial marker system usually consists of markers, a detection algorithm, and a coding system.
Existing detection algorithms are hand-crafted based on traditional image processing techniques such as edge detection, blob detection, and image binarization.
Due to these limited low-level techniques, only simple marker appearance can be supported.
Examples of popular marker families are shown in Fig.~\ref{fig:example_markers}.
ARToolkitPlus~\cite{Wagner2007}, AprilTag~\cite{olson2011apriltag, wang2016apriltag}, and ArUco~\cite{garrido2014automatic, Garrido-Jurado2015} use black-and-white checkerboard markers whose quadrilateral boundary is detected by analyzing lines.
The bits are decoded by binarizing small regions. 
However, marker pose estimated with four corners is less accurate under noise or motion blur.
Recent marker systems further introduce redundant keypoints defined with structures inside the marker.
Specifically, TopoTag~\cite{yu2020topotag} limits the arrangement of bits in a checkerboard layout and provides fixed regions.
Each region represents a keypoint and a bit.
RuneTag~\cite{bergamasco2011rune, bergamasco2016an} arranges solid dots on concentric circles.
The dots' presence or absence represents bits, and their centroids are used for pose estimation.
Both TopoTag and RuneTag require good image quality such that the smallest inner elements can be accurately estimated.

On the other hand, the simplicity of marker appearance complicates the design of coding system.
Specifically, the arrangement of black-and-white cells in a checkerboard marker should be rotationally asymmetric to avoid pose ambiguity.
To reduce the possibility of confusion with environmental elements, a high number of bit transitions is also encouraged~\cite{garrido2014automatic, Garrido-Jurado2015}.
The above rules also apply to RuneTag.
Moreover, as bits may be decoded incorrectly, intermarker distance is often enlarged by further introducing Hamming distance, which in turn limits the marker library compacity.

To alleviate the limitations of existing marker systems, we present a new viewpoint for marker design and detection.
As illustrated in Fig.~\ref{fig:general_marker_design}, a marker is a set of keypoints attached with different local patterns representing digital symbols.
Instead of hand-crafting shape analysis methods, we propose a general framework \emph{DeepTag} which directly estimates keypoints and digital symbols from an image utilizing end-to-end convolutional neural networks (CNN).
DeepTag also supports existing marker families. 
Compared to existing detection algorithms, DeepTag introduces a much larger number of keypoints utilizing the structures inside a marker to improve pose accuracy.
For example, all the cell centroids of a checkerboard marker are regarded as keypoints (Fig.~\ref{fig:example_marker_kpts}(d)).
Keypoints can also be defined on both visible and invisible dots (see Fig.~\ref{fig:example_marker_kpts}(g) for RuneTag).
By utilizing subpixel information, markers can still be detected even when edge detection and image binarization are intractable (Fig.~\ref{fig:detect_cube}).
Moreover, DeepTag makes it possible to design new markers with customized local patterns.
As shown in Fig.~\ref{fig:example_revised_markers}(a), new patterns can be applied to an existing coding system.
Complex patterns not only reduce the possibility of confusion with environmental elements but also provide local structures for accurate keypoint detection.
Furthermore, by applying more than two categories of patterns, rotational asymmetricity is no longer a requirement for the coding system (Fig.~\ref{fig:example_revised_markers}(b) and Fig.~\ref{fig:example_revised_markers}(c)).

\REVISED{
The E2ETag~\cite{peace2020e2etag} and the concurrent work DeepFormableTag~\cite{yaldiz2021deepformabletag} utilize CNN to improve marker detection.
They automatically generate markers consist of free-form textures or colors which are optimized for detection and ID determination.
But dedicated printer and camera are necessary during application, and users' flexibility to design preferred patterns is also restricted.
Moreover, they directly regress the marker corners or 6-DoF, resulting in poor marker pose accuracy, which limits single-marker applications.
In contrast, our framework allows user to adopt existing or self-designed markers, and achieves high pose accuracy for single marker by employing dense keypoints.
}

CNN-based keypoint detection has been successfully applied to tasks like human pose estimation~\cite{cao2019openpose} and face alignment~\cite{bulat2017far}.
In these applications, an object consists of $n$ keypoints that belong to $n$ different categories which can be uniquely determined with local context (e.g., an eye can be determined without the visibility of the mouth).
In contrast, the local context on a fiducial marker only determines the keypoint location and the encoded digital symbol.
Thus, we propose to first regress unordered keypoints and then sort them to determine the marker pose and ID.
Similar to existing CNN-based techniques, another major challenge is training data collection.
Real images or videos include neither marker pose nor keypoint positions, and it is intractable to accurately annotate dense keypoints.
Instead, we propose to synthesize training data for existing and newly-designed marker families.
To bridge the gap between synthesized data and real scenarios, our generated images are augmented with image degradation that is far more challenging than real applications.
Existing image synthesis techniques often rely on sophisticated photorealistic rendering, and images are generated and stored in advance~\cite{simon2017hand, varol2017learning}.
In contrast, we generate training data on the fly based on basic image processing techniques.
To validate DeepTag and the existing methods, besides the existing dataset, we further collect a large dataset of 70k images containing markers placed in different view distances and angles.
Experiments show that DeepTag well supports both existing and newly-designed markers.
Moreover, DeepTag outperforms the existing hand-crafted methods in terms of both detection robustness and pose accuracy.
In summary, the contributions of DeepTag include:

\begin{itemize}
	\item We propose DeepTag, a general framework for fiducial marker design and detection. 
	\item DeepTag supports a large variety of existing marker families and greatly outperforms the existing methods in terms of both detection robustness and pose accuracy.
	\item DeepTag can be easily applied to new marker families with customized local patterns.
	\item We propose an effective procedure to synthesize training data without manual annotation, which facilitates broader applications of DeepTag.
\end{itemize}

\section{Related Work}
\subsection{Fiducial Marker System}

The state-of-the-art fiducial marker systems usually consist of a coding/decoding system, a set of marker images, an algorithm to detect and validate the markers.
Square and circular patterns are commonly used in fiducial markers, as they are easy to print and detect.

Gatrell et al.~\cite{Gatrell1991} use concentric contrasting circles (CCC) for fiducial marker design.
Correspondences between the centroids of the found ellipses are used to validate the marker existence.
This design is slightly improved in~\cite{Cho1998} by drawing the circles with different colors and multiple scales.
Dedicated "data rings" are added in the marker design in ~\cite{Knyaz1998, Naimark2002}.
Fourier tags~\cite{Sattar2007, Xu2011} encode multi-bit information in the fequency domain.
RuneTag~\cite{bergamasco2011rune, bergamasco2016an} uses rings of dots to represent cyclic codes and provide a large number of keypoints for robust pose estimation.

Matrix~\cite{Rekimoto1998}, CyberCode~\cite{Rekimoto2000} and VisualCode~\cite{Rohs2004} are the early proposals of square markers.
The well known ARToolkit~\cite{Kato1999} includes a fixed square border for pose estimation.
Validation is based on correlation between the observed image with the prototype images.
Due to the unreliability of image correlation, it appears difficult to offer a large and reliable marker library.
ARTag~\cite{Fiala2005, Fiala2010} and ARToolkitPlus~\cite{Wagner2007} use digital coding systems instead of analog systems to represent marker IDs.
Robust digital coding techniques are used to achieve higher performance.
The digital symbols are represented with black-and-white cells, thus the shape analysis is convenient and digital symbols are decoded via image binarization.
Based on the basic checkerboard marker design, BinARyID~\cite{Flohr2007} attempts to generate a rotationally asymmetric coding system.
AprilTag~\cite{olson2011apriltag, wang2016apriltag} presents a faster and more robust detection algorithm for ARTag.
ArUco~\cite{garrido2014automatic, Garrido-Jurado2015} attempts to generate coding systems with large intermarker distance and a high number of bit transitions.
Romero-Ramirez et al.~\cite{Romero-Ramirez2018} propose a speeded-up checkerboard marker detection algorithm for video sequences.
ChromaTag~\cite{Degol2017} incorporates color into AprilTag to improve detection speed.
TopoTag~\cite{yu2020topotag} presents a checkerboard marker design with fixed regions, where the type of center cells of the regions are used to represent bits. It achieves higher pose accuracy by offering more keypoints instead of using four corners alone.

The proposed DeepTag is a general framework that supports detection of a large variety of existing marker families.
Instead of relying on edge detection and image binarization, DeepTag directly regresses keypoints and digital symbols from local shapes with CNN.
Compared to traditional algorithms, a larger number of keypoints can be defined and estimated to achieve higher pose accuracy.
By utilizing subpixel information, detection robustness is also dramatically increased.
As it is not necessary to hand-craft shape analysis, DeepTag also supports newly-designed marker families with customized local patterns.

\begin{figure}[tp]
	\centering
	\includegraphics[width=0.4\textwidth]{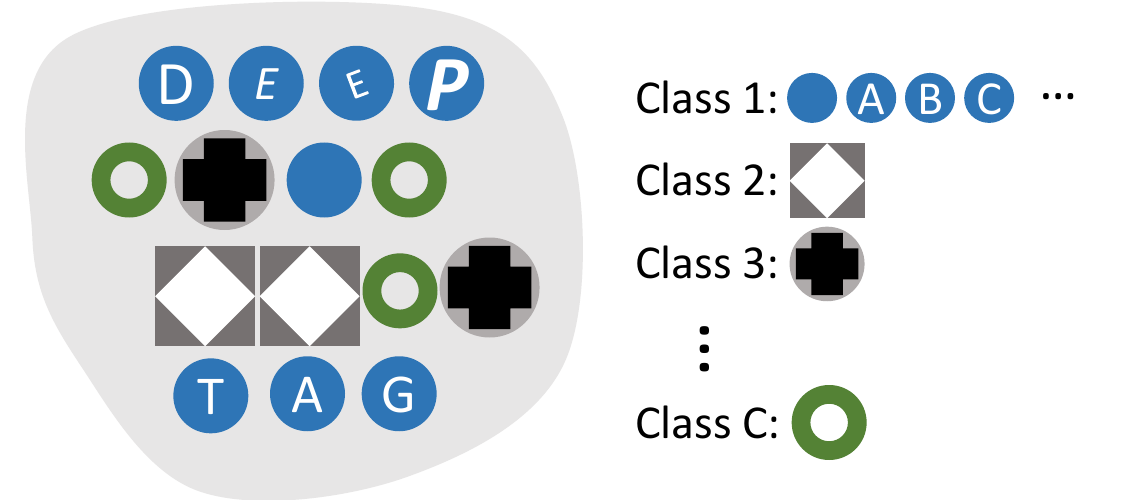}
	\caption{A general marker design supported by DeepTag. It supports using $C$ categories of local patterns that represent $C$ possible digital symbols. }
	\label{fig:general_marker_design}
	\vspace{-0.05in}
\end{figure}

\vspace{-0.1in}
\subsection{Convolutional Neural Network}
CNN has been successfully applied to many computer vision tasks. 
Our goal targets to detect the fiducial markers from the scene and estimate the keypoints of the markers.
Among the recent CNN-based applications, the most relevant ones are object detection and keypoint detection.

Object detection techniques estimate the location of an object while at the same time determine its category. 
Recently, a variety of CNN-based detection schemes have been proposed for object detection.
Fast R-CNN~\cite{girshick2015fast} and Faster R-CNN~\cite{ren2015faster} first generate a large number of region proposals, and then classify them and predict the bounding boxes.
Single Shot Multibox Detector (SSD)~\cite{liu2016ssd} uses a single convolutional network to predict the bounding boxes and their class probabilities.
Without resampling pixels or features for bounding box hypotheses, SSD achieves a much faster computational speed than  Fast R-CNN and Faster R-CNN.
In our proposed method, SSD is used to generate marker hypotheses for keypoint detection.

Keypoint detection is a basic task in applications like human pose estimation~\cite{cao2019openpose} and face alignment~\cite{bulat2017far}.
The keypoints are then used for alignment between images and virtual models.
The total number of keypoints on a single object is predefined and required for the network architecture.
The index of a keypoint, which is also the class label, is uniquely determined by local context.
By contrast, the keypoints on a fiducial marker are categorized by the digital symbols encoded with local shapes or colors.
The index of a keypoint is determined only after all the keypoints are sorted with their locations.

Object counting is another slightly related application.
Crowd counting~\cite{wang2015deep, boominathan2016crowdnet, onoro2016towards} concerns the total number of crowded objects instead of their locations.
Cell nucleus detection~\cite{tofighi2019prior} estimates the locations of small repeated shapes.
Similarly, our task detects and counts small unordered shapes.
But by contrast, for the purpose of subsequent pose estimation, the accurate locations of the shapes should further be sorted and estimated.

Deep ChArUco~\cite{hu2019deep} utilizes CNN to detect keypoints of a ChArUco board which is a chessboard of ArUco markers.
It outperforms traditional methods under poor lighting or extreme motion blur.
But this method can only detect one chessboard where the marker family and IDs are fixed.
E2ETag~\cite{peace2020e2etag} utilizes CNN to generate markers with random textures and detect them in an end-to-end manner.
But its library only contains a very limited number (i.e., 30) of markers, and incorrect ID identification happens when the image quality is not good.
The lack of regular boundary or local structures inside a marker may lead to inaccurate marker pose.
\REVISED{
Similarly, the concurrent work DeepFormableTag~\cite{yaldiz2021deepformabletag} also utilizes CNN for marker generation and recognition.
ID decoding and marker location are robust under various conditions. 
But marker ID representation relys on automatically generated free-form colors information, which requires dedicated printer to accurately print the marker and high-quality camera to capture the colors.
Similar to many traditional methods, it determines marker pose with only four corners, which leads to poor pose accuracy and limits single-marker scenario.    
}
In contrast to these works, DeepTag is a general CNN-based fiducial marker detector that supports a wide variety of existing and newly-designed markers.
Accurate 6-DoF pose and marker ID can be estimated even under very poor image quality.

DeepTag accomplishes a new task that is different from the above-mentioned detection techniques.
First, multiple artificial objects are detected from the image.
Then, for each object, small unordered shapes are accurately detected, categorized, and sorted.
For the existing works, developing a CNN-based detection technique often requires a large annotated dataset.
However, accurate annotation requires special equipment~\cite{simon2017hand}, and it is intractable to annotate dense keypoints on fiducial markers.
Instead, we present an effective procedure to generate training data without human annotation.
Purely trained with synthetic data, DeepTag successfully generalizes to real-life applications.


\section{Overview}
\subsection{Marker Design}
An abstracted marker supported by DeepTag is illustrated in Fig.~\ref{fig:general_marker_design}.
It consists of local patterns whose centroids and categories represent keypoints and digital symbols.
Similar local patterns are classified into a single category, and patterns of different categories should be distinguishable in terms of shapes or colors.
By regarding black-and-white cells or dots as local patterns, this marker abstraction directly supports existing popular markers shown in Fig.~\ref{fig:example_markers}.

\begin{figure}[tp]
	\centering
	\includegraphics[width=0.5\textwidth]{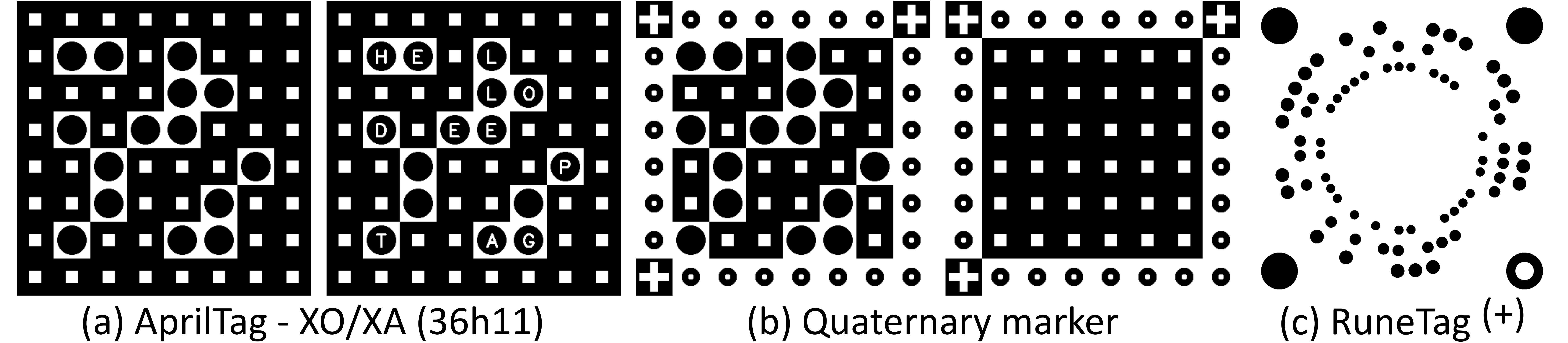}
	\caption{Examples of new marker design supported by DeepTag. (a) AprilTag with new patterns. (b) Markers with four different patterns. (c) RuneTag with four more keypoints.}
	\label{fig:example_revised_markers}
\end{figure}

This general abstraction also provides a good flexibility to design new markers with customized local patterns.
For example, based on bits arranged in a $N\times N$ grid, two categories of local shapes can be adopted to create new markers (Fig.~\ref{fig:example_revised_markers}(a)).
The uniqueness of the patterns greatly decreases the possibility of confusion with environment elements.
For existing binary checkerboard markers (e.g., AprilTag, ArUco, and ARToolkitPlus), the same target can only be achieved by introducing a high number of bit transitions.
Besides two categories of patterns that represent bits, extra categories can be used to indicate border and orientation.
Thus rotation asymmetricity is no longer a requirement for the coding system (e.g., bits of the second marker in Fig.~\ref{fig:example_revised_markers}(b) can be all "0"s).
Similarly, four extra keypoints can be used to facilitate determining rotation of RuneTag (Fig.~\ref{fig:example_revised_markers}(c)). 

\begin{figure*}[tp]
	\centering
	\includegraphics[width=\textwidth]{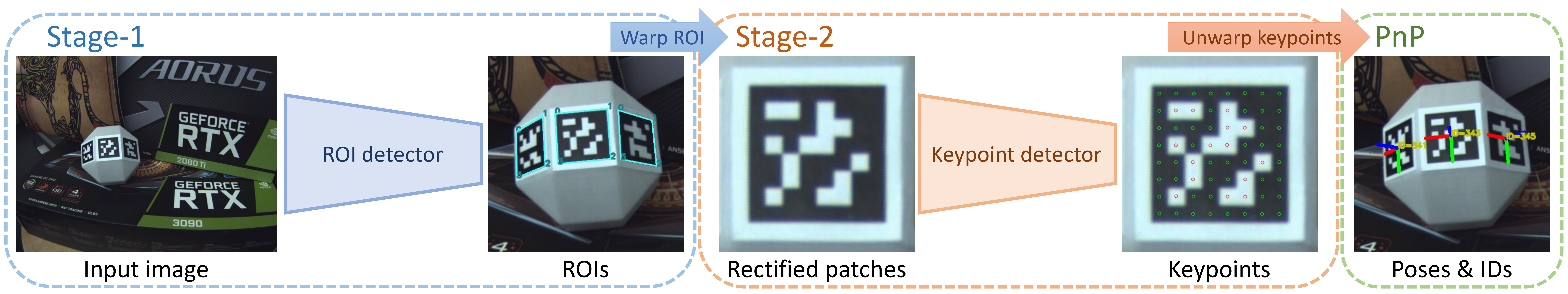}
	\caption{System overview. A two-stage scheme is adopted to detect marker ROIs and estimate keypoints for each ROI. Marker 6-DoF poses and ID information can then be determined.}
	\label{fig:overview}
\end{figure*}

\subsection{Keypoint Definition}
\begin{figure}[tp]
	\centering
	\includegraphics[width=0.5\textwidth]{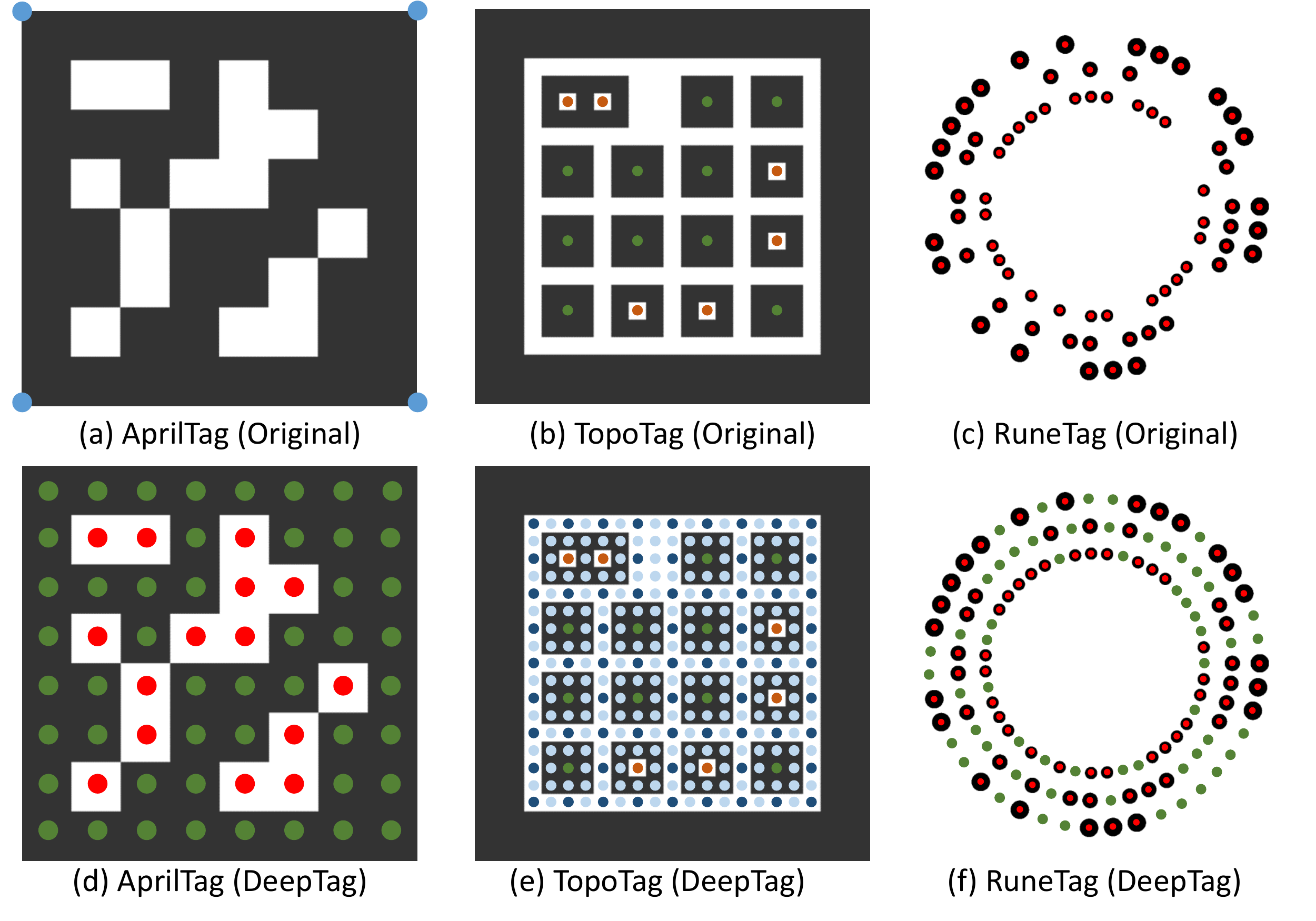}
	\caption{Keypoint definition for existing markers. Compared with the original definition, DeepTag utilizes keypoints that are dense and regularly arranged. Green, red, and blue dots are three types of keypoints with "0", "1" and non-encoding bit. (a-c) keypoint definitions of the original methods. (d-f) keypoint definitions in DeepTag.}
	\label{fig:example_marker_kpts}
\end{figure}

Existing keypoint definition suffers from low keypoint capacity or irregular arrangement.
Specifically, keypoints on a checkerboard marker are originally defined only on four corners of its boundary and encode no bits (Fig.~\ref{fig:example_marker_kpts}(a)).
A small number of keypoints lead to inaccurate estimated pose when the image quality is not good.
On the other hand, irregular keypoint arrangement on TopoTag (Fig.~\ref{fig:example_marker_kpts}(b)) and RuneTag (Fig.~\ref{fig:example_marker_kpts}(c)) makes it difficult to sort the keypoints.

In contrast, DeepTag defines keypoints in a dense and regular way for all the existing markers.
A larger number of keypoints improve pose accuracy, and regular arrangement facilitates keypoint sorting.
Specifically, keypoints on a checkerboard marker are defined on centroids of all the $N\times N$ cells, each of which is attached with a bit (Fig.~\ref{fig:example_marker_kpts}(d)).
TopoTag is a specially designed binary checkerboard marker with fixed regions. 
Besides regarding it as a common binary marker, another smart option is classifying the keypoints into three categories (Fig.~\ref{fig:example_marker_kpts}(e)).
Two of them represent "0"s and "1"s, and the other one doesn't encode any bit. 
It is also reasonable to define the keypoints only on every other cell (i.e., exclude light blue dots in Fig.~\ref{fig:example_marker_kpts}(e)) without affecting all the encoding bits.
For RuneTag, keypoints are originally defined only on the visible dots (Fig.~\ref{fig:example_marker_kpts}(c)), and the total number of keypoints varies among different markers.
Instead, we define keypoints on both the visible and invisible dots (Fig.~\ref{fig:example_marker_kpts}(g)) to utilize the regular arrangement, which makes the sorting scheme much easier.

It's worth noting that keypoint definition is independent from the adopted patterns. For example, newly-designed AprilTag (Fig.~\ref{fig:example_revised_markers}(a)) and the original AprilTag (Fig.~\ref{fig:example_markers}(a)) share the same keypoint definition, while newly-designed quaternary markers (Fig.~\ref{fig:example_revised_markers}(b)) also include $N\times N$ keypoints but with four different class labels.


\begin{figure}[tp]
	\centering
	\includegraphics[width=0.5\textwidth]{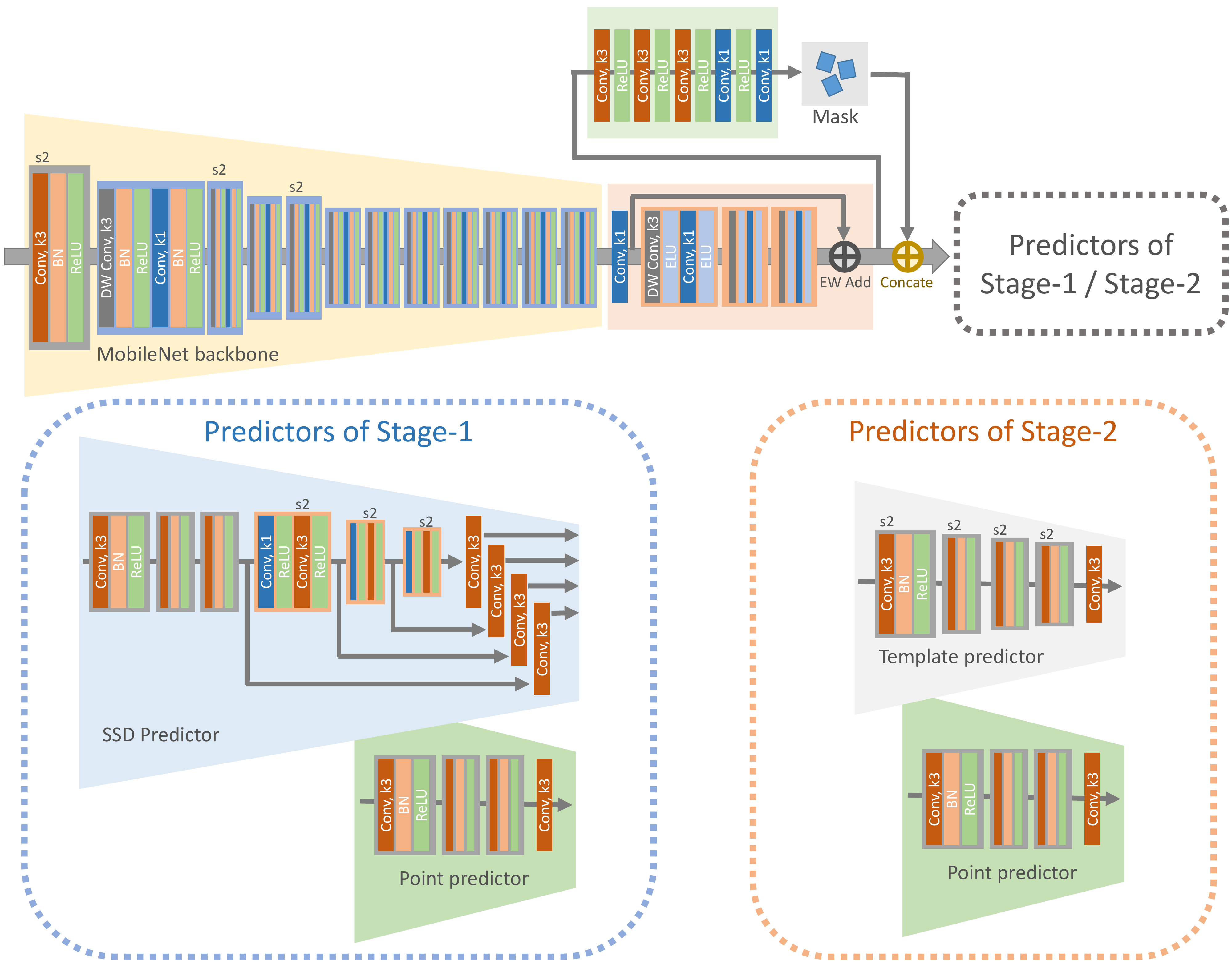}
	\caption{Network structure for ROI and keypoint detection. "Conv" and "DW Conv" denote convolutional layer and depthwise convolutional layer respectively, "k1" and "k3" for spatial kernels of size 1 and 3, "s2" for convolution with a stride of 2, and "EW Add" for element-wise addition.}
	\label{fig:keypoint_network}
\end{figure}

\subsection{Detection Pipeline}
Given an input image, DeepTag detects groups of keypoints, and each group belongs to a marker.
To achieve robustness and accuracy, similar to~\cite{simon2017hand}, DeepTag adopts a two-stage detection scheme illustrated in Fig.~\ref{fig:overview}, with detailed network structure in Fig.~\ref{fig:keypoint_network}.
Region of Interest (ROI) is first estimated for each potential marker, and each ROI is represented with at least four non-collinear points arranged in clockwise order.
Then, a rectified patch is generated with a homography matrix that maps the ROI to predefined points in the patch (e.g., a four-point ROI is mapped to a fixed square).
In the patch, keypoints and digital symbols are estimated and sorted.
Marker ID is acquired by checking the marker library with the decoded digital symbols.
Keypoint positions in the original image are obtained by applying the inverse homography matrix.
With a known physical marker size, the marker’s 6-DoF pose is estimated with the Perspective-n-Point (PnP) algorithm.
This two-stage framework can be applied to different marker families by defining ROI, keypoints, and digital symbol encoding scheme accordingly.

\section{Approach}
\subsection{ROI Detection}
\label{sec:roi_detect}
Existing methods usually detect potental markers in a bottom-up manner~\cite{Fiala2005, Fiala2010, Wagner2007, olson2011apriltag, wang2016apriltag,yu2020topotag}.
Edges or blobs are detected with traditional low-level image processing techniques and then grouped to be marker ROIs.
Differently, DeepTag detects marker ROIs in a top-down manner by directly determining their existence along with at least four sorted non-colinear points.
We regard it as a standard multi-scale object detection task and utilize SSD~\cite{liu2016ssd} to regress bounding boxes along with the auxiliary information.

Bounding box $B=(l, cls)$ is a axis-aligned minimum rectangle enclosing a marker, where the localization $l$ is presented with the center $(l_{cx}, l_{cy})$ and the size $(l_{w}, l_{h})$, and $cls^{(i)} \in \{1,\dots, C\}$ is the class that the object belongs to.
When there is only one type of marker, the total number of classes $C$ is 2 (marker and background). 
In the form of a center and four sorted corner points, $ROI^{(b)} = (P_0, P_1, P_2, P_3, P_4)$ is predicted along with the bounding box $b$.

\begin{figure}[tp]
	\centering
	\includegraphics[width=0.5\textwidth]{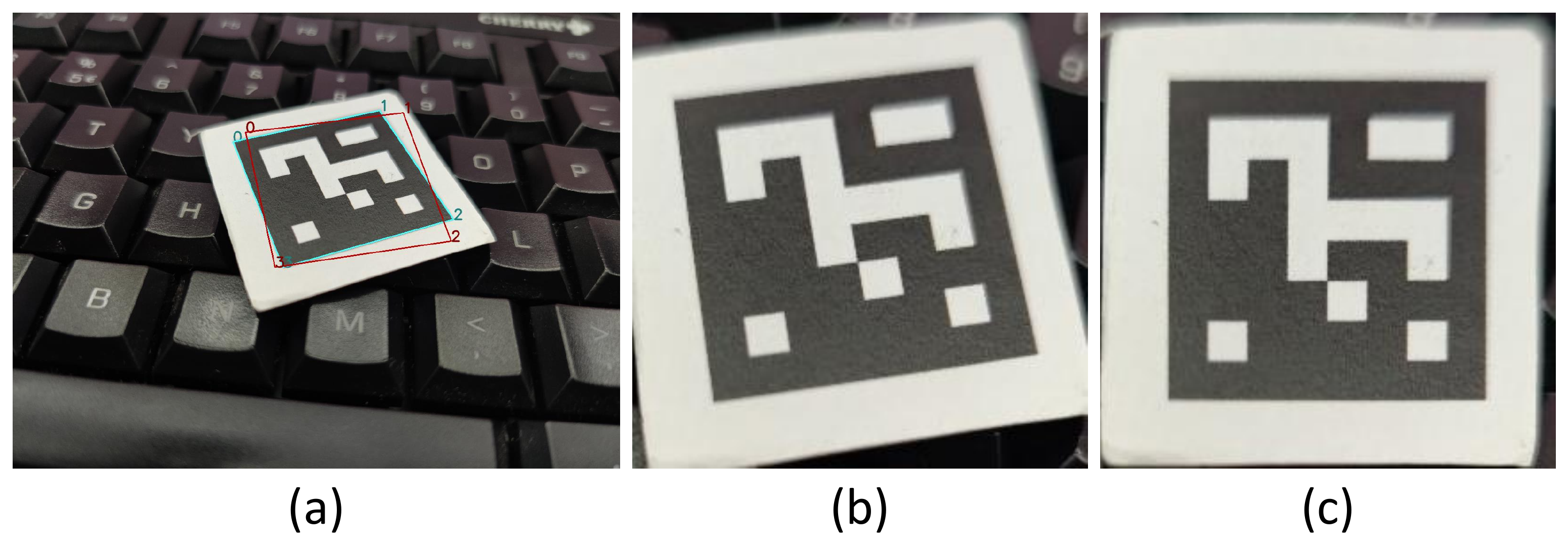}
	\caption{ROI prediction and refinement. (a) Initial ROI (red) and refined ROI (cyan). (b) Patch warped from (a) with initial ROI. (c) Patch warped from (a) with refined ROI.}
	\label{fig:roi_prediction}
\end{figure}

Note that, $ROI^{(b)}$ will be inaccurate for markers in a large view angle (Fig.~\ref{fig:roi_prediction}(a-b)).
For further ROI refinement, unordered corners $\{\hat{P}_k\}$ are detected and assigned to the markers.
$\{\hat{P}_k\}$ are regarded as one-scale objects without size and only based on local context.
To correctly assign $\hat{P}$ to a marker, the directional vector $\hat{V}$ that points from $\hat{P}$ to the marker center is also precited.
By refining $ROI^{(b)}$ with $\{\hat{P}_k\}$, the final $ROI^{(r)}$ is much more accurate (Fig.~\ref{fig:roi_prediction}(c)).

\textbf{Network structure.}
As shown in Fig.~\ref{fig:keypoint_network}, given an input image $I$ of size $h\times w$, features $X_{0}$ with spatial dimension $h/8 \times w/8$ are extracted with a MobileNet~\cite{howard2017mobilenets} backbone encoder and a residual block.
MobileNet is more efficient than VGG-16~\cite{simonyan2014very} because it uses depthwise separable convolution to decrease the computational complexity while preserving the capacity of feature extraction.
As fiducial markers consist of small simple shapes, we propose to extract intermediate middle-level features before high-level object detection.
A 2-channel mask $M$ of size $h/8 \times w/8 \times 2$ that represents the segmentation of the markers and background is predicted by processing $X_{0}$ with several convolutional layers.
Then the backbone features $X_{0}$ and the mask $M$ are concatenated to be the intermediate features $X_1$.
Two predictors are used to predict bounding boxes and corner points respectively.

An SSD predictor estimates $\{B_i\}$ and $\{ROI^{(b)}_i\}$ based on predefined multi-scale default bounding boxes.
Following \cite{liu2016ssd}, the output localization $\tilde{l}$ is the offset from the default bounding box $l^{(d)}$ to the predicted localization $l$:
\begin{equation}
	\tilde{l} = \bigg(\frac{l_{cx}-l^{(d)}_{cx}}{l^{(d)}_{w}}, \frac{l_{cy}-l^{(d)}_{cy}}{l^{(d)}_{h}}, \log\Big( \frac{l_{w}}{l^{(d)}_{w}}\Big), \log\Big( \frac{l_{h}}{l^{(d)}_{h}}\Big)\bigg).
\end{equation}
Based on the predicted bounding box localization $l$, we define $ROI^{(b)} = (P_0, P_1, P_2, P_3, P_4)$ as relative coordinates: 
\begin{equation}
\tilde{P}_j^{(b)} = \Big(\frac{P_{j,x} -l_{cx}}{l_{w}} - 0.5,\frac{P_{j,y}  -l_{cy}}{l_{h}} - 0.5 \Big).
\label{eqn:relative_coord}
\end{equation}

Another one-scale predictor estimates $\{\hat{P}_k\}$ and $\{\hat{V}_k\}$ based on predefined points arranged in a $h/8 \times w/8$ grid. 
The output localization $\tilde{P}$ is the offset from the default point $P^{(d)}$ to the predicted localization $\hat{P}$:
\begin{equation}
	\tilde{P} = \big(\hat{P}_{x}-P^{(d)}_{x}, \hat{P}_{y}-P^{(d)}_{y}\big).
	\label{eqn:corner_coord}
\end{equation}
For numeric stability, a normalized 2D vector $\hat{V}$ is represented in a redundant form:
\begin{equation}
	\tilde{V} = \Big(\frac{\hat{V}_x + 1}{2}, \frac{\hat{V}_y + 1}{2}, 1-\frac{\hat{V}_x + 1}{2}, 1-\frac{\hat{V}_y + 1}{2}\Big).
\end{equation}

\textbf{Loss functions.}
During training, SSD uses a matching strategy to assign ground truth bounding boxes to the predefined anchors.
A similar matching strategy is also adopted on the one-scale predictor for corner points.
There are two categories of anchors: positive anchors which correspond to ground truth bounding boxes or corners, and negative anchors that are background.
The loss of classification confidence is defined based on both positive and negative anchors, while the loss for information of bounding boxes or corners is defined only on positive anchors.
Let $x^p_{ij} = \{0,1\}$ be an indicator for matching the $i$-th default box/point to the $j$-th ground truth box/corner of category $p$.
The confidence loss is the softmax loss over multiple class confidences $c$:
\begin{equation}
	\begin{aligned}
    L_{conf}(c)=& -\sum_{i\in Pos}{x^p_{ij}\log\Big(\frac{\text{exp}
	(c^p_i)}{\sum_p\text{exp}(c^p_i)}\Big)} \\
	&-\sum_{i\in Neg}{\log\Big(\frac{\text{exp}(c^0)}{\sum_p\text{exp}(c^p_i)}\Big)}.
	\end{aligned}
    \label{eqn:loss_conf}   
\end{equation}
The loss for location and other auxiliary information is a Smooth L1 loss~\cite{girshick2015fast} between the predicted data $g$ with the ground truth data $g^*$:
\begin{eqnarray}
    L_{info}(g, g^*) = \sum_{i\in Pos}^N{x^p_{ij} \text{smooth}_{L1}\big(g_i-g^*_j\big)}
    \label{eqn:loss_smooth_l1}
\end{eqnarray}
Based on Eqn.~\ref{eqn:loss_conf} and Eqn.~\ref{eqn:loss_smooth_l1}, the losses for bounding box and for corners are defined as:
\begin{equation}
    L_{box} = \frac{1}{N}\Big(L_{conf}\big(c^{(b)}\big) + L_{info}\big((\tilde{l}, \{\tilde{P}^{(b)}_j\}) , (\tilde{l}, \{\tilde{P}^{(b)}_j\})^*\big)\Big)  ,
\end{equation}
\begin{equation}
L_{corner} = \frac{1}{N}\Big(L_{conf}\big(c^{(p)}\big) + L_{info}\big( (\tilde{P}, \tilde{V}), (\tilde{P}, \tilde{V})^*\big)\Big),
\label{eqn:accurate_corner}
\end{equation}
where $N$ is the total number of the matched bounding boxes or corners, and the loss is set to $0$ when $N=0$.
$c^{(b)}$ and $c^{(p)}$ are class confidences of bounding boxes and unordered corner points respectively.
The mask loss is an L2 loss between the predicted masks $M$ with the ground truth masks $\hat{M}$:
\begin{eqnarray}
    L_{mask} = ||M-\hat{M}||^2_2.
	\label{eqn:loss_mask}
\end{eqnarray}
The overall loss function of Stage-1 is the combination of losses for two predictors and the intermediate mask:
\begin{eqnarray}
    L_{ROI} =  L_{box} + L_{corner} + L_{mask}.
\end{eqnarray}

\begin{figure}[tp]
	\centering
	\includegraphics[width=0.4\textwidth]{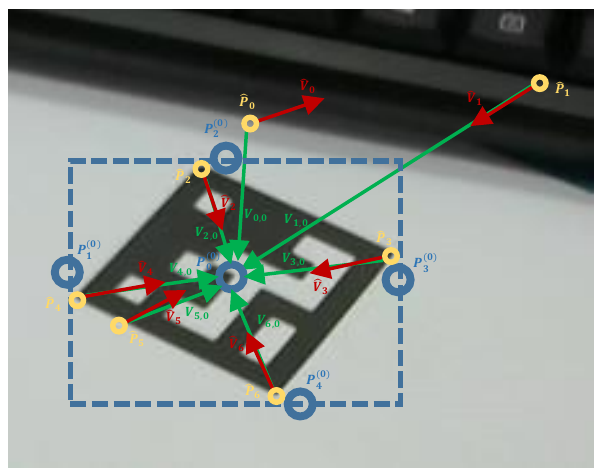}
	\caption{RIO refinement.}
	\label{fig:assign_corners}
\end{figure}

\textbf{Postprocessing.}
One marker may be predicted with multiple bounding box anchors.
Instead of using non-maximum-suppression (NMS) to remove redundant bounding boxes, we filter the outputs with their confidence scores.
Then the final output bounding boxes are acquired by clustering the filtered network outputs with their localizations $l$.
With this clustering procedure, the predicted ROIs encounter less jitter.
Similarly, the predicted unordered corners are also clustered with not only their locations $\hat{P}$ but also the corresponding directions $\hat{V}$.
For corners of nearby markers with similar $\hat{P}$ but different $\hat{V}$, they would be assigned to different clusters.

To acquire the final $ROI^{(r)}$ from $ROI^{(b)}$, we adopt two criterions to assign unordered corners $\{\hat{P}_k\}$ to markers:
\begin{equation}
	Dist(\hat{P}_k, l) = \max\bigg(\Big|\frac{\hat{P}_{k,x} -l_{cx}}{0.5\cdot l_{w}} - 1\Big|, \Big|\frac{\hat{P}_{k,y} -l_{cy}}{0.5\cdot l_{h}} - 1\Big|\bigg)
\end{equation}
\begin{equation}
    \cos(\hat{V}_k, V_{k,i})= \frac{\hat{V}_k \cdot V_{k,i} }{||\hat{V}_k|| \ ||V_{k,i}||}
\end{equation}
where $Dist(\hat{P}_k, l)$ calculates the relative coordinates of $\hat{P}_k$ with the respect to the bounding box localization $l$, 
and $Dist(\hat{P}_k, l)\in [0,1]$  if $\hat{P}_k$ is inside the bounding box.
$\cos(\hat{V}_k, V_{k,i})$ checks the angle between the estimated direction $\hat{V}_k$ and the real direction $V_{k,i}$ from the $k$-th corner to the center of the $i$-th marker. 
The $k$-th corner is assigned to the $i$-th bounding box only if $Dist(\hat{P}_k, l_i) < 1.5$ and $\cos(\hat{V}_k, V_{k,i})>0.6$.
In the example in Fig.~\ref{fig:assign_corners}, $\hat{P}_0$ and $\hat{P}_1$ are discarded because $\cos(\hat{V}_0, V_{0,0})$ and $Dist(\hat{P}_1, l_0)$ do not fullfill the requirements.
Then, from $\{\hat{P}_2,\hat{P}_3,\hat{P}_4,\hat{P}_5,\hat{P}_6\}$, the nearest points of $P^{0}_1$, $P^{0}_2$, $P^{0}_3$, and $P^{0}_4$ are selected, resulting the final $ROI^{(r)}_0 =(\hat{P}_4,\hat{P}_2,\hat{P}_3,\hat{P}_6)$.

\begin{figure}[tp]
	\centering
	\includegraphics[width=0.5\textwidth]{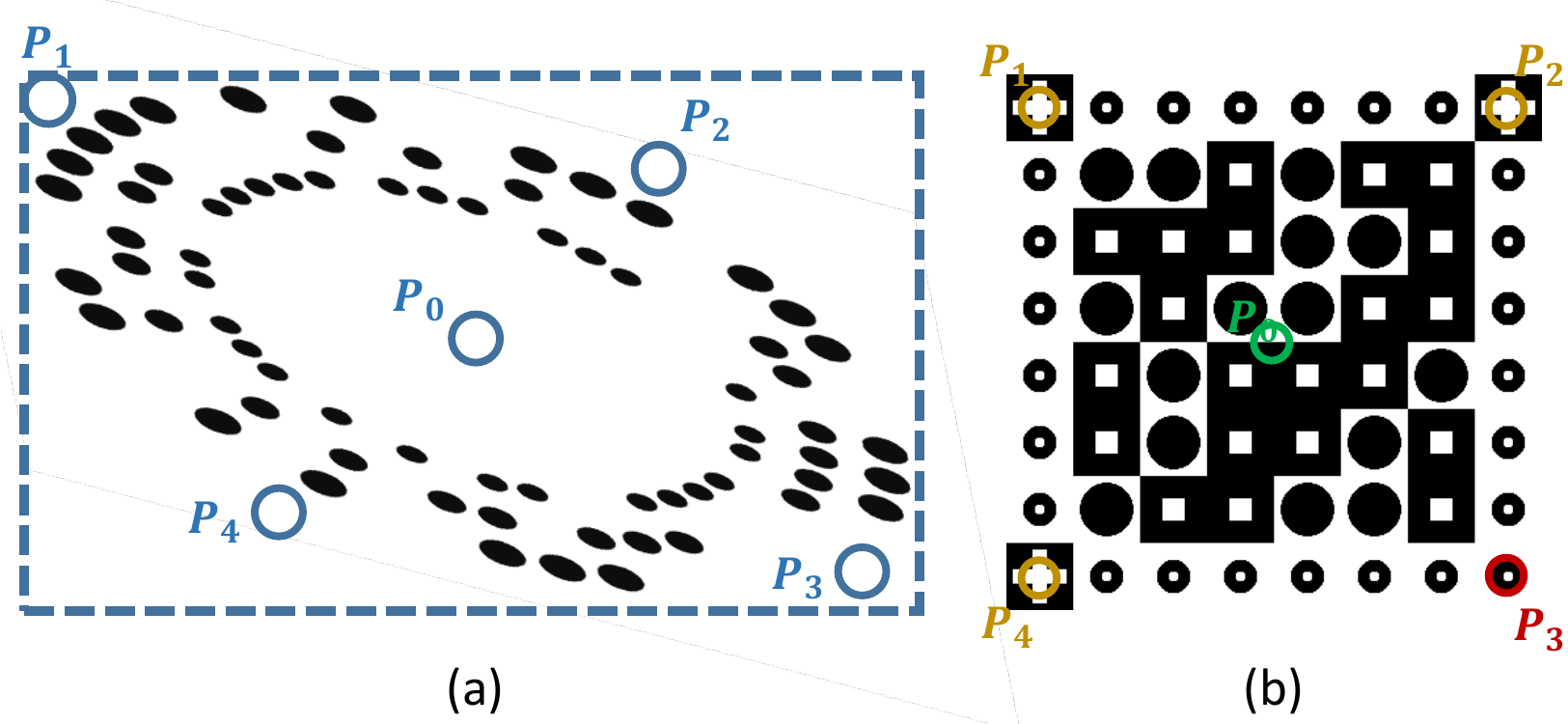}
	\caption{ROI definition. (a) ROI definition for RuneTag with vertices of the minor and major axis of the ellipse. (b) ROI definition for one of our newly-designed marker with four keypoints of two classes where $P_3$ is with a different class label against $\{P_1, P_2, P_4\}$.}
	\label{fig:roi_extension}
\end{figure}

\textbf{ROI of other marker families.} 
Other marker families can be supported by modifying the ROI definition.
Specifically, for RuneTag without explicit border, ROI can be defined with four vertices of the ellipse and can be predicted along with the bounding box (Fig.~\ref{fig:roi_extension}(a)).
On the other hand, ROI can also be defined with specially designed keypoints instead of vertices of the boundary (Fig.~\ref{fig:roi_extension}(b)).
It is also possible to determine the marker orientation if points of ROI belong to multiple categories.

\subsection{Keypoint Detection}
\label{sec:landmark_decode}
With an ROI predicted by the previous stage, a patch is warped from the image by mapping the ROI to a predefined template.
Specifically, the four-point ROI template is a square $T^{(ROI)} = ((a, a), (1-a, a), (1-a, 1-a), (a, 1-a))$, where $a = 0.15$ in our implementation.
Then unordered keypoints positions and digital symbols $\{(K_j, t_j)\}$ are directly regressed on the rectified patch.
For pose estimation and marker ID determination, $\{(K_j, t_j)\}$ are sorted to an ordered sequence according to their localizations.

\textbf{Keypoint sorting.}
For marker families with regular keypoint arrangement, an ordered keypoint template $T^{(K)} = (K^{(T)} _0 , K^{(T)} _1 , \dots)$ can be acquired with $T^{(ROI)}$.
Specifically, $T^{(K)}$ is an $N\times N$ grid for checkerboard markers with $N$ determined by counting $\{K_j\}$, and then the nearest keypoint of each $K^{(T)}$ is selected from $\{K_j\}$.
Note that this sorting procedure may fail for marker families without stable ROI (e.g., RuneTag).
Instead of hand-crafted rules, we further regress an ROI template $\hat{T}^{(ROI)} = (P^{(T)} _0 , P^{(T)} _1 , \dots)$ which leads to a refined keypoint template $\hat{T}^{(K)}$.

\textbf{Network structure and loss functions.}
The network structures are shown in Fig.~\ref{fig:keypoint_network} with two predictors to estimate $\{(K_j, t_j)\}$ and $\hat{T}^{(ROI)}$ respectively.
Similar to detecting accurate corners $\{\hat{P}_j\}$ in the previous stage, we adopt a one-scale anchor-based predictor to estimate $\{(K_j, t_j)\}$.
Given an input image of size $w\times w$, the anchor points are arranged in a $w/8\times w/8$ grid.
On the other hand, a template predictor takes intermediate features $X_1$ as input, and estimates $\hat{T}^{(ROI)}$ in a one-scale manner with $2\times2$ anchors.

Based on Eqn.~\ref{eqn:loss_conf} and Eqn.~\ref{eqn:loss_smooth_l1}, the loss functions for two predictors are defined as:
\begin{eqnarray}
	L_{keypoint} = \frac{1}{N}\Big(L_{conf}\big(c^{(K)}\big) + L_{info}\big( \tilde{K}, \tilde{K}^*\big)\Big) ,\\
	L_{template} = \frac{1}{N}\Big(L_{conf}\big(c^{(T)}\big) + L_{info}\big( \tilde{P}^{(T)}, \tilde{P}^{(T)*}\big)\Big) ,
\end{eqnarray}
where $\tilde{K}$ and $\tilde{P^{(T)}}$ are localization calculated from $K$ and $P^{(T)}$, and based on Eqn.~\ref{eqn:corner_coord}, 
$c^{(K)}$ and $c^{(T)}$ are class confidences.
The final loss for keypoint detection includes the above two losses and a mask loss defined by Eqn.~\ref{eqn:loss_mask}:
\begin{eqnarray}
    L_{detect} =  L_{keypoint} + L_{template} + L_{mask}.
\end{eqnarray}
Similar to the previous stage, the final $\{(K_j, t_j)\}$ is acquired by clustering the network output.
On the other hand, postprocessing is not necessary for $\hat{T}^{(ROI)}$ as it is already in a sorted form.


\subsection{Training Data Preparation}
\label{sec:training_data_prepare}
Training images are synthesized by adding markers to different real scenarios.
Background images are randomly sampled from the indoor dataset~\cite{quattoni2009recognizing} and  OpenImages~\cite{openimages}.
The former dataset simulates the real usage environments, while the latter includes artificial objects where markers are likely added.
We use objects with basic shapes that are similar to markers, including book, laptop, tablet computer, computer monitor, keyboard, headphone, mobile phone, etc.

\begin{figure}[tp]
	\centering
	\includegraphics[width=0.5\textwidth]{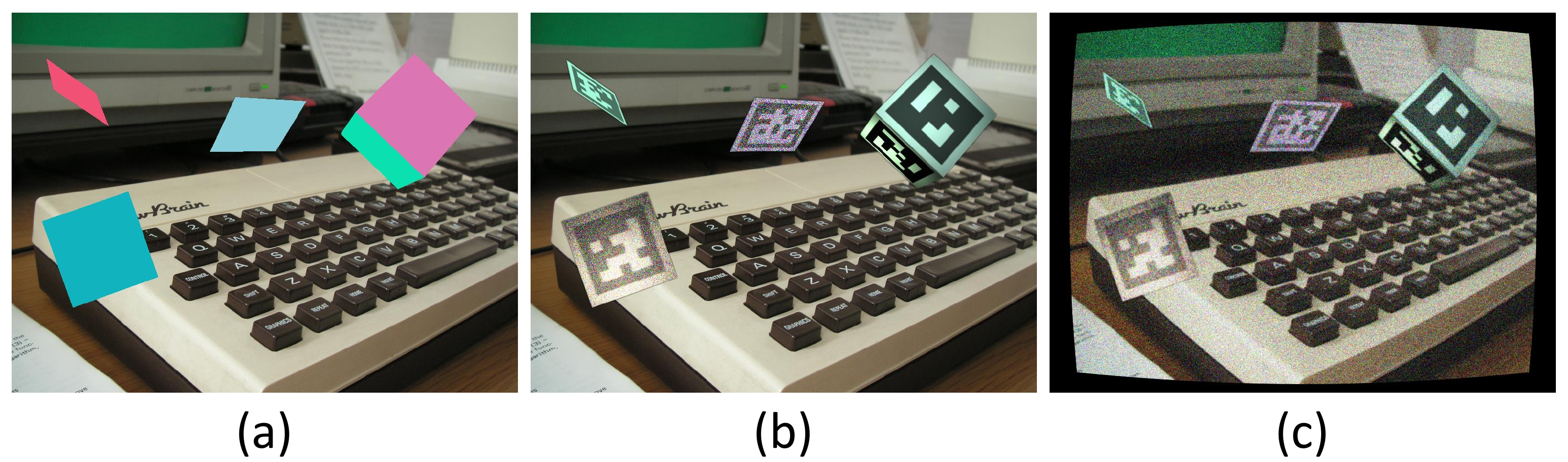}
	\caption{Data synthesis. (a) Random marker locations. (b) Adding the markers. (c) Final synthetic image.}
	\label{fig:data_synthesis}
\end{figure}

Binarized markers are randomly sampled from several predefined marker families.
Then, random grayscale, spotlighting effect, Gaussian blur, Gaussian noise and motion blur are added to simulate various printing conditions, lighting conditions and marker movements.
Augmented markers are added into random locations with quadrilateral boundaries (Fig.~\ref{fig:data_synthesis}(a-b)).
Finally, the whole image is further augmented with Gaussian blur, Motion blur, random white balance, Gaussian noise and barrel distortion to simulate different image capturing processes (Fig.~\ref{fig:data_synthesis}(c)).

With the training data generated on the fly, we separately train the networks of Stage-1 and Stage-2 using Adam solver~\cite{kingma2014adam} with a batch size of 4.
The networks are trained for 100K iterations with a learning rate of 3e-4.
Stage-1 is trained with images of size $768\times 768$. 
When training Stage-2, markers in a synthesized image are cropped and rectified to patches of size $256\times 256$ by mapping the ground truth ROI to a random quadrangle patch $((a_1, a_2), (1-a_3, a_4), (1-a_5, 1-a_6), (a_7, 1-a_8))$, where $a_i$ is a random value uniformly sampled in $[0.05, 0.25]$.

\section{Experiment}
\label{sec:experiment}

\subsection{Datasets and Validation Metrics}
The two basic targets during marker detection are determining the existence of markers and estimating their poses.
We validate both for the proposed and existing detection algorithms in terms of detection robustness and pose accuracy.
Detection robustness is measured with recall and precision of detection.
Pose accuracy is evaluated with the error between the estimated pose with the ground truth.

\textbf{Yu's dataset.} 
By using a programmable robot arm to control the ground truth camera movements, 
Yu et al.~\cite{yu2020topotag} collected a large dataset that includes marker families shown in Fig.~\ref{fig:example_markers}.
For markers with multiple library settings like ArUco and AprilTag, each of them will be captured.
There are 10 static camera poses with 9 grouth truth pose offsets. 
Under a certain camera pose, around 775 images are captured for each marker to calculate the average pose and jitter.



\textbf{Our new dataset.}  
In Yu's dataset, the camera moves without rotation, where markers are kept almost parallel to the camera plane.
This makes the dataset may not well simulate the complex real-life use scenarios, where markers or cameras rotate freely and images with various blur in different distances and angles.
To fill this gap, we further collect a new large dataset by placing the markers in combinations of different distances and angles.
Following Yu's dataset, four representative checkerboard markers are selected, i.e., AprilTag (36h11), ARToolKitPlus, ArUco (36h12) and TopoTag.
We further evaluate three newly-designed markers (Fig.~\ref{fig:example_revised_markers}(a, c) that are supported by DeepTag.
Specifically, AprilTag-XO (36h11) and AprilTag-XA (36h11) contain different newly-designed local patterns with bits arranged as AprilTag (36h11).
RuneTag$^{(+)}$ is a modified version of RuneTag that is still compatible with the original detection algorithm.

\begin{figure}[tp]
	\centering
	\includegraphics[width=0.5\textwidth]{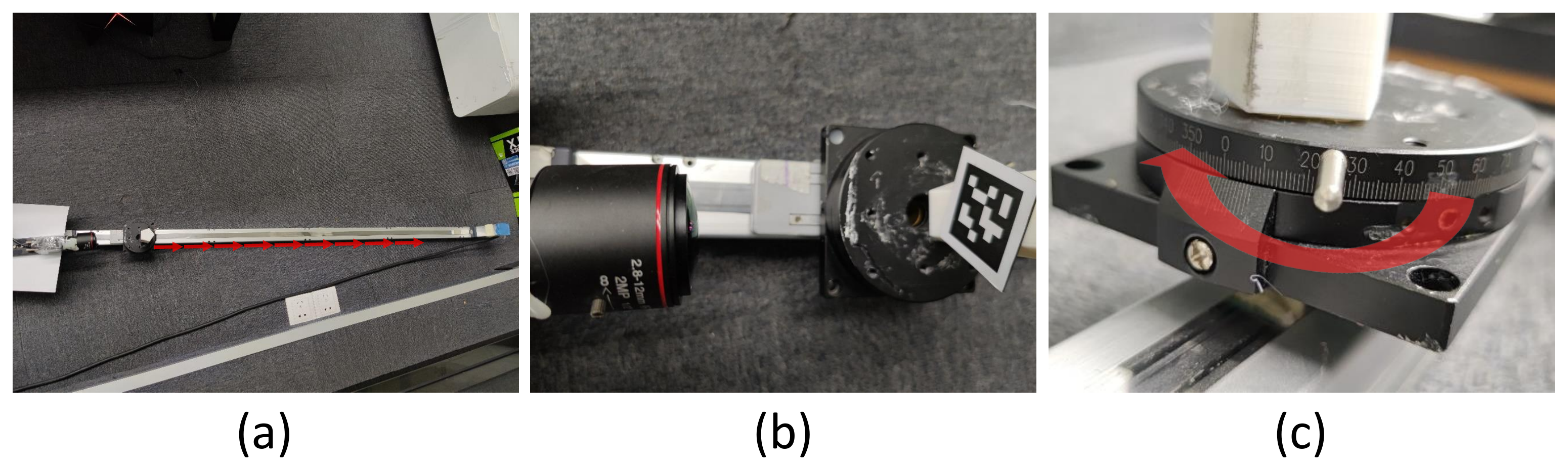}
	\caption{Device setup for our new dataset. (a) Markers move along a linear axis away from the camera (red arrows). (b) A close view of the camera and marker. (c) The platform that rotates the marker along an axis.}
	\label{fig:linseq_device_setting}
    \vspace{-0.2in}
\end{figure}

We use an industrial global shutter camera with $98^{\circ}$ diagonal field of view.
Images are captured at 38.8fps of size $1280\times 960$ with a fixed exposure time of 20ms.
As shown in Fig.~\ref{fig:linseq_device_setting}, a rotational flatform is used to control the view angle, and a 1D linear programmable moving platform\footnote{We use a 1D linear programmable moving platform from IAI (RCP5-BA4). Link: \href{https://www.intelligentactuator.com/RCP5-BA4/}{https://www.intelligentactuator.com/RCP5-BA4/}} is used to accurately move the marker.
Markers move away from the camera with a constant moving step of 10cm, with each round in one of in total 10 different view angles ($0^{\circ}$, $10^{\circ}$, $20^{\circ}$, $\dots$, $60^{\circ}$, $65^{\circ}$, $70^{\circ}$, $75^{\circ}$).
The nearest and farthest distances are 10cm and 100cm respectively.
There are in total 90 moving offsets that server as ground truth with moving distance of 10cm and zero rotation.
To evaluate pose error and jitter separately, 100 images are captured for each pose, resulting totally 10k images for each marker.
As shown in Fig.~\ref{fig:linseq_image_examples}, the markers are intentionally placed out of alignment with image axes to simulate real use scenarios. For better analysis, we further split the dataset into two subsets, i.e., "Normal" set with markers $\leqslant 50$cm and $\leqslant 60^{\circ}$ and "Difficult" set with markers $\textgreater 50$cm or $\textgreater 60^{\circ}$. 

\textbf{Validation metrics.}
Among all the detected markers returned by an algorithm, correct ones are defined as true positives (TP), and the ones with incorrect location or ID are defined as false positives (FP).
False negatives (FN) are defined as any marker that exists in the image but is not detected by the algorithm.
Precision and recall are defined as $\frac{TP}{TP+FP}$ and $\frac{TP}{TP+FN}$ respectively.
Following \cite{Degol2017}, correct marker detections are defined with both correct marker location and ID.
Marker location is correct if there is at least 50\% intersection over the union between the detection with the ground truth.
Both pose accuracy and jitter are measured in terms of rotation and translation. As mentioned earlier, pose accuracy is evaluated with the ground truth pose offset between two adjacent poses.
Pose jitter is evaluated using the standard deviation (STD) metric.


\subsection{Ablation Study}
\label{sec:ablation_study}
Stage-1 predicts ROIs with two possible versions, i.e., with or without refinement.
By iteratively calling Stage-2 multiple times, ROIs can be further updated with the keypoints predicted by the previous iteration.
We evaluate the accuracy of the ROIs and the final marker pose to compare different settings on three checkerboard markers, i.e., ARToolkitPlus, ArUco and AprilTag.

\begin{figure}[tp]
	\centering
	\includegraphics[width=0.5\textwidth]{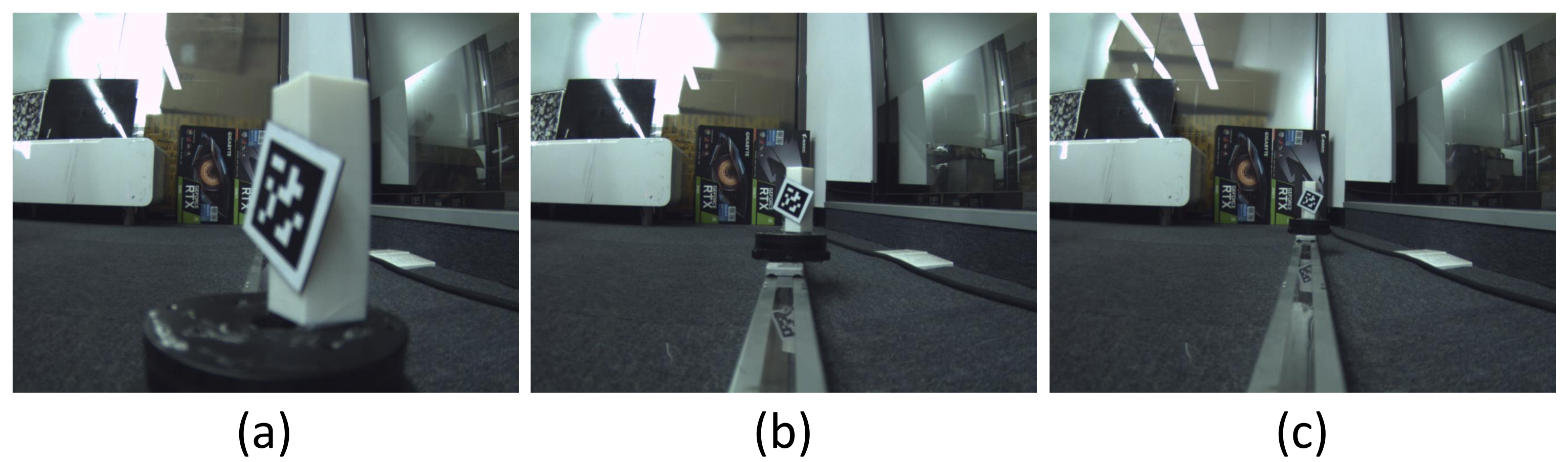}
	\caption{Example images in our new dataset, where markers are placed in different distances and angles. (a) $10$cm and $60^{\circ}$. (a) $30$cm and $30^{\circ}$. (a) $60$cm and $0^{\circ}$.}
	\label{fig:linseq_image_examples}
    \vspace{-0.2in}    
\end{figure}

\begin{figure*}[tp]
	\centering
	\includegraphics[width=\textwidth]{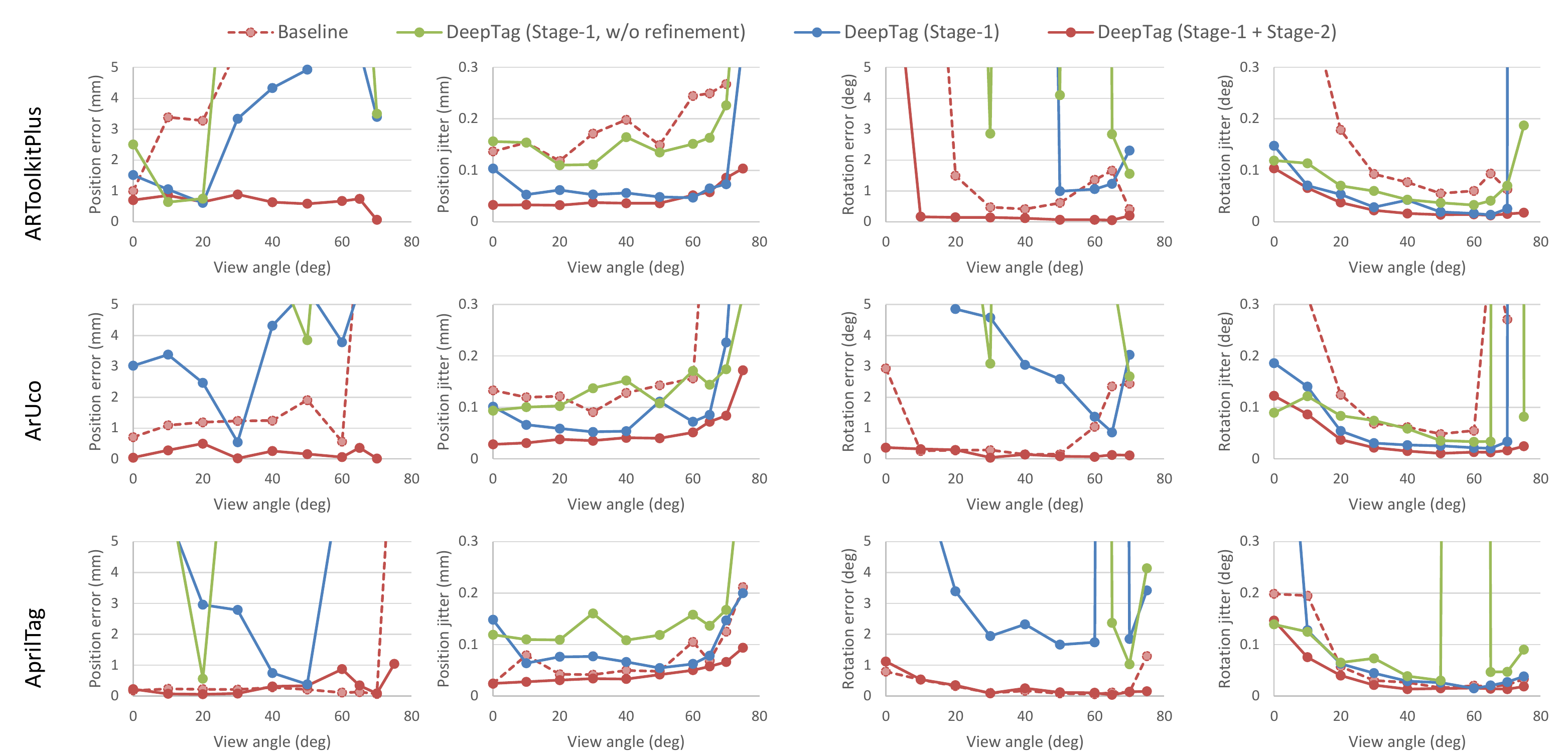}
	\caption{Ablation study of different ROI versions. ROI from Stage-1 with refinement is significantly improved over the non-refinement version. The ROI updated by Stage-2 further improves and outperforms ”baseline”s for all markers.}
	\label{fig:linseq_accuracy_plot_ablation}
\end{figure*}

Due to the lack of ground truth annotations, 
we evaluate the ROIs by evaluating the marker pose estimated using the four keypoints of ROIs.
Among three tested ROI versions, two are predicted by Stage-1 (with or without refinement) while another one is updated by the first iteration of Stage-2.
Evaluation is conducted in a fixed view distance of 30cm (to 40cm) with different view angles using our new dataset, see sample images in Fig.~\ref{fig:linseq_image_examples}(b). 
Results are shown in Fig.~\ref{fig:linseq_accuracy_plot_ablation} where the original detection algorithms are "baseline". Results show that refinement in Stage-1 leads to significant improvement over the non-refinement version.
The ROI updated by Stage-2 further improves from Stage-1 ROIs and outperforms "baseline" for all markers.

\begin{figure}[tp]
	\centering
	\includegraphics[width=0.5\textwidth]{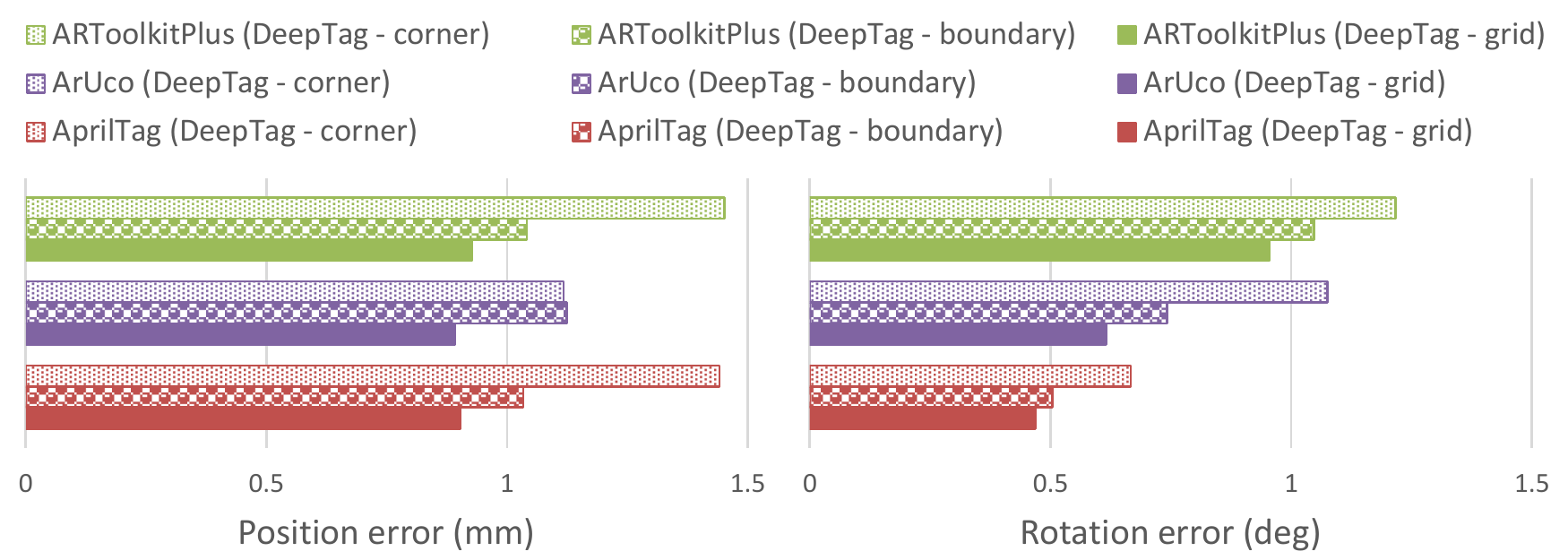}
	\caption{\REVISED{Ablation study of different numbers of keypoints. Using all the keypoints consistently achieves the best results.}}
	\label{fig:linseq_accuracy_plot_ablation_kpts_num}
\end{figure}

\begin{figure}[tp]
	\centering
	\includegraphics[width=0.5\textwidth]{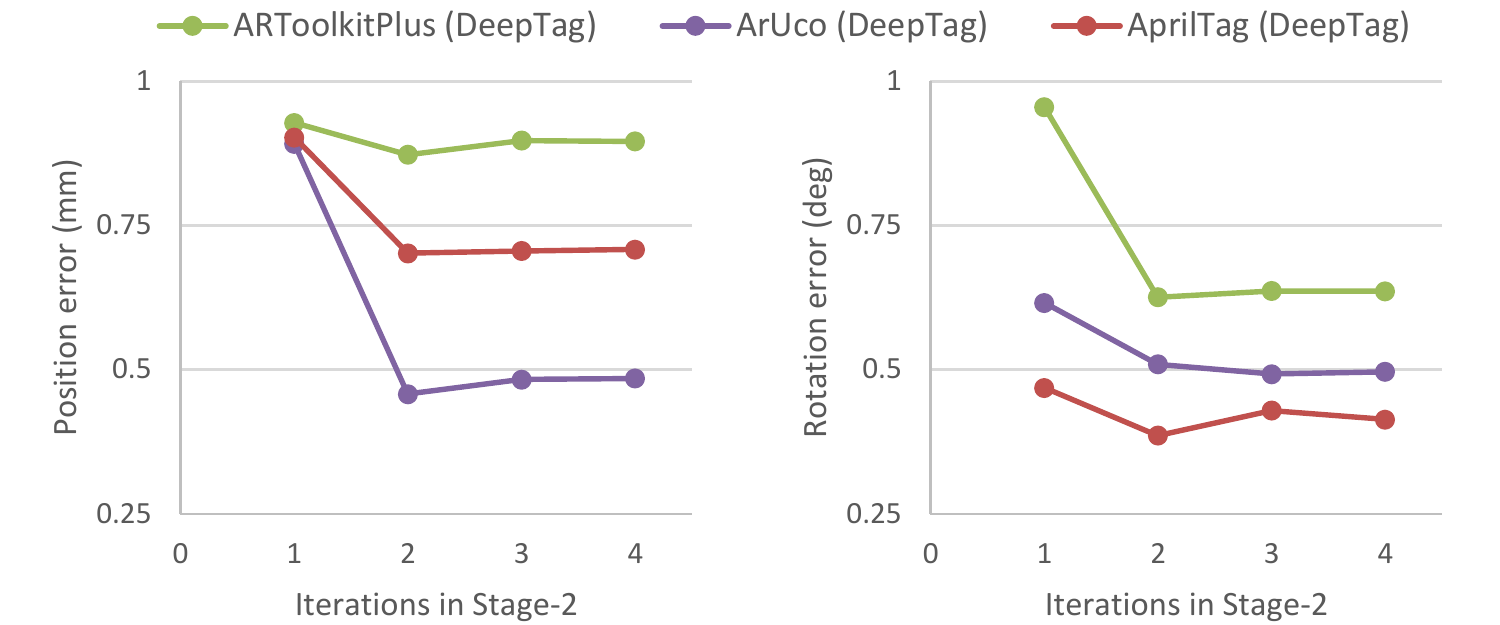}
	\caption{Ablation study of different Stage-2 iterations. The results become stable after two iterations.}
	\label{fig:linseq_accuracy_plot_ablation_iteration}
\end{figure}

\REVISED{
Performance comparisons of Stage-2 with different keypoint settings are shown in Fig.~\ref{fig:linseq_accuracy_plot_ablation_kpts_num}.
For a marker with keypoints in a $N \times N$ grid, 'corner', 'boundary', and 'grid' respectively denote pose estimation in Stage-2 with 4 corners, $4N-4$ boundary points, and the whole $N \times N$ grid.
From the results, it is obvious that pose estimation with more keypoints generally achieves better position and rotation accuracy, and the best results are consistently achieved by using all the keypoints.
Furthermore, performance comparisons of Stage-2 with all keypoints but}
with different number of iterations are shown in \autoref{fig:linseq_accuracy_plot_ablation_iteration}.
From the results, it is obvious that pose accuracy generally improves from the first to the second iteration and becomes stable afterward.

According to the above study, for detection on still images, the suggested setting is the combination of Stage-1 with refined ROI and Stage-2 \REVISED{with all the keypoints and} with two iterations. We fix this setup for all the following experiments.

\subsection{Detection Robustness}
\begin{figure}[tp]
	\centering
    \includegraphics[width=0.5\textwidth]{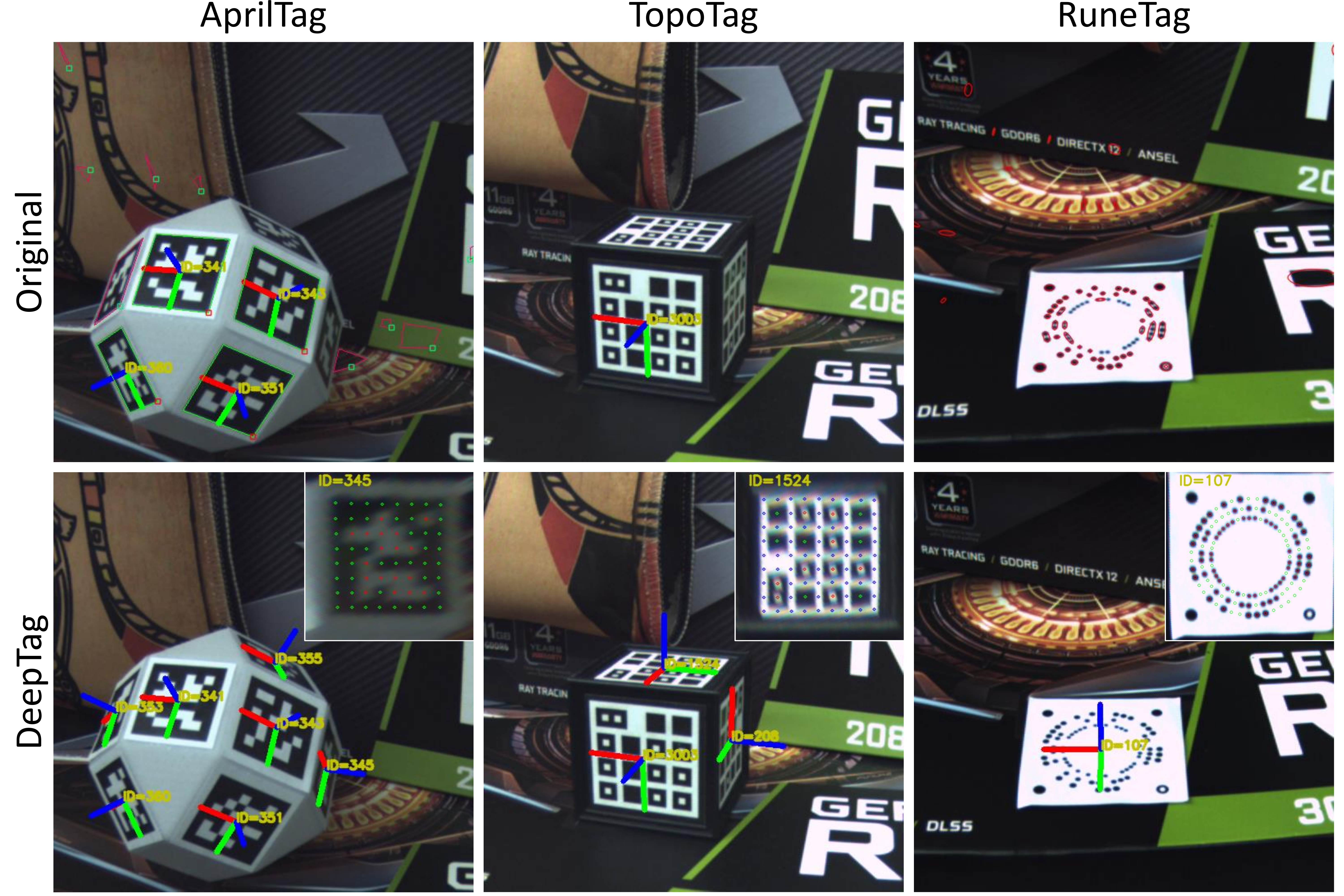}
	
	\caption{Qualitative detection results. DeepTag successfully detects all markers while original methods either miss some or totally fail.}
	\label{fig:detect_cube} 
\end{figure}

Fig.~\ref{fig:detect_cube} shows the qualitative detection results of DeepTag vs. existing methods.
Both AprilTag and TopoTag fail in large view angles, and RuneTag fails to detect the marker because the black dots cannot be successfully located.
In contrast, DeepTag successfully detects all markers in these challenging cases. 
As further illustrated in the rectified patches, keypoints and the embedded digital symbols are also accurately estimated by DeepTag.

Detection robustness is further quantitatively evaluated on both Yu's dataset and our new dataset, and the results are summarized in Tab.~\ref{table:topotag_dataset} and Tab.~\ref{table:linseq_dataset} respectively.
Compared with existing methods, DeepTag consistently achieves equal or higher recall and precision for all markers on both datasets. 
In Yu's dataset, DeepTag achieves perfect 100\% recall and precision for all markers. For our new dataset, especially for  "Difficult" set, DeepTag achieves $17\mathtt{\sim}1055\%$ higher in recall over original methods, which further validates DeepTag's robustness in challenging use scenarios.

As shown in Tab.~\ref{table:linseq_dataset}, RuneTag and TopoTag fail when view distance or view angle becomes larger as they require high image quality to detect the complex local structures such as dots or the smallest cell in the checkerboard layout.
In contrast, DeepTag utilizes subpixel information to detect and achieves a much better detection recall.
For our newly-designed AprilTag-XO and AprilTag-XA that contain complex local patterns, DeepTag can still achieve comparable performance compared to the simpler markers.

\textbf{Noise and motion blur.}
To simulate low-light scenarios, different levels of noise and motion blur are synthesized and added to images.
For each marker, 20 images are selected from our new dataset. 
10 of them contain markers of 30cm and 30\textdegree \ (median distance and angle in "Normal" set, see Fig.~\ref{fig:linseq_image_examples}(b) for image samples), while the other 10 images are with the same view angle but in the next location (i.e., 40cm).
Gaussian noise of different standard deviations $\sigma$ or linear motion blur of different kernel sizes $k$ are added to the original images.
When applying linear motion blur, 10 evenly sampled moving directions are considered.
Note that, RuneTag is not included in this validation as the original method fails even for the original images. 

\begin{figure}[tp]
	\centering
    \includegraphics[width=0.5\textwidth]{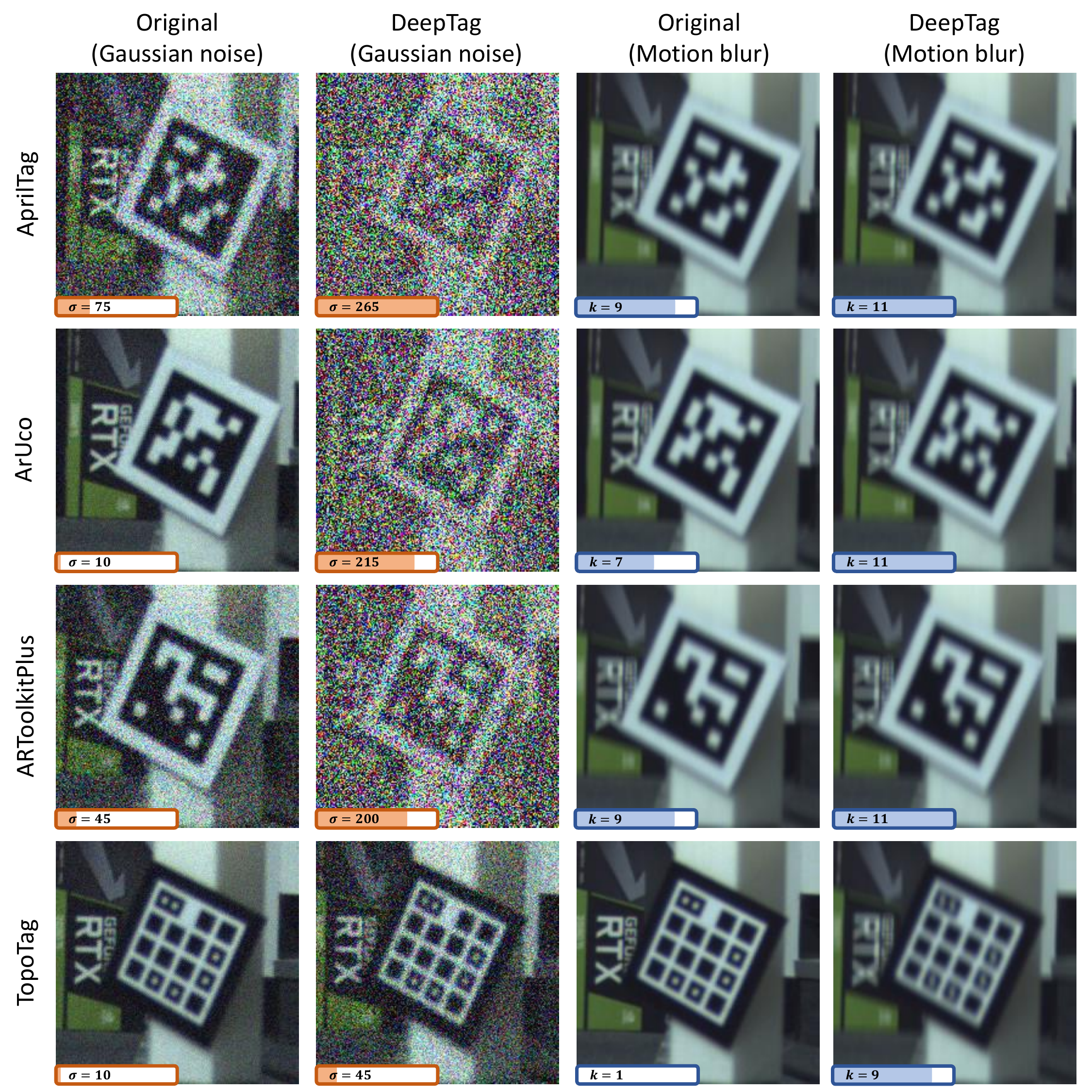}
	
	\caption{Cropped image patches with the most severe level of Gaussian noise or motion blur that each method can tolerate. DeepTag consistently tolerates more severe Gaussian noise and motion blur than all the original methods.}
	\label{fig:gnoise_mblur_max_level} 
\end{figure}

Fig.~\ref{fig:gnoise_mblur_max_level} shows the cropped patches with the most severe level of Gaussian noise or motion blur that each method can tolerate. DeepTag consistently tolerates more severe Gaussian noise and motion blur than all the original methods.
Specifically, AprilTag, ArUco and ARToolkitPlus can be robustly detected by DeepTag under Gaussian noise with $\sigma = 200$ or motion blur with $k = 11$.
For TopoTag that contains smaller cells, DeepTag can still deal with Gaussian noise with $\sigma = 45$ or motion blur with $k = 9$.

\subsection{Pose Accuracy}
\label{sec:pose_accuracy}

The overall performance of pose accuracy on both datasets are summarized in Tab.~\ref{table:topotag_dataset} and Tab.~\ref{table:linseq_dataset}. 
DeepTag consistently achieves the best results across all markers for all the four metrics (i.e., position error, position jitter, rotation error and rotation jitter) on both datasets. 
Specifically, for our new dataset, DeepTag achieves $54\mathtt{\sim} 70\%$ pose accuracy improvement for the "Normal" set by comparing the best results of DeepTag with the best of existing methods across all markers. This improvement further increases to $83\mathtt{\sim} 95\%$ for the "Difficult" set.
It's also interesting to find that even the worst results of DeepTag are still much better with $39\mathtt{\sim} 73\%$ pose accuracy improvement than the best of existing methods, which verifies DeepTag's pose superiority on challenging use scenarios.

\begin{table*}[tp]
    \centering
    \caption{Pose results on Yu's Dataset~\cite{yu2020topotag}. Best results are shown in bold and underlined.}
    \vspace{-0.1in}
    \begin{tabular}{c|c|cc|cc|cc}
    \hline \hline
    \multirow{2}{*}{Method} & \multirow{2}{*}{Marker} &\multicolumn{2}{c}{Position (mm)}&  \multicolumn{2}{|c}{Rotation (deg)} &  \multicolumn{2}{|c}{Detection (\%)} \\ \cline{3-8}
    & & Error&  Jitter & Error&  Jitter   &  Recall & Precision   \\ \hline
    &AprilTag (16h5)                 & 2.894 &0.079 & 0.031 &0.654 &77.29 & 99.88 \\ 
    &AprilTag (25h7)                 & 2.704 &0.104 & 0.026 & 0.879 &75.71 & 100\\ 
    &AprilTag (25h9)                 & 3.178 &0.087 &0.024 &0.673 &80.41 & 100\\ 
    &AprilTag (36h9)                 & 3.228 &0.102 &0.024 &0.753 & 78.70 & 100\\ 
    Original&AprilTag (36h11)       & 1.402 & 0.074 &\topone{0.010} &  0.133 & 100 & 99.99 \\ 
    &ARToolKitPlus                   & 8.923 &1.134 & 0.040 & 0.421 & 98.30 & 100 \\ 
    &ArUco (16h3)                    & 8.191 & 0.363 & 0.248 & 0.230 & 100 & 99.91 \\ 
    &ArUco (25h7)                    & 10.049 & 0.364 & 0.225 & 0.322 & 99.01 & 100 \\ 
    &ArUco (36h12)                   & 8.768 & 0.573 & 0.078 & 0.526 & 99.47 & 100 \\ 
    &TopoTag                         & 1.011 & 0.055 & 0.019 &  0.058 & 100 & 100\\ 
    &RuneTag              & - & -&- & -& 0 & 0\\ \hline
    &AprilTag (16h5)        & 0.805 & 0.043 & 0.023 & 0.061 & 100 & 100 \\ 
    &AprilTag (25h7)        & 0.645 & 0.036 & 0.021 & 0.066 & 100 & 100\\ 
    &AprilTag (25h9)        & 0.968 & 0.045 & 0.017 & 0.012 & 100 & 100\\ 
    &AprilTag (36h9)        & 1.099 & 0.034 & 0.016 & 0.011 &100 & 100\\ 
    &AprilTag (36h11)       & \topone{0.609} & 0.034 & 0.015 & 0.010 & 100 & 100 \\ 
    DeepTag&ARToolKitPlus   & 0.718 & 0.033 & \topone{0.010} & \topone{0.009} & 100 & 100 \\ 
    &ArUco (16h3)           & 0.643 & 0.043 & 0.016 & 0.064 & 100 & 100\\ 
    &ArUco (25h7)           & 0.802 & 0.042 & 0.023 & 0.010 & 100 & 100 \\ 
    &ArUco (36h12)          & 0.846 & 0.037 & 0.015 & 0.062 & 100& 100 \\ 
    &TopoTag                & 0.875 & \topone{0.023} &0.023 &  0.024 & 100 & 100\\ 
    &RuneTag                & 0.904 & 0.040 & 0.015 &  0.086 & 100 & 100\\ 

    \hline\hline
    \end{tabular}
    \label{table:topotag_dataset}
\end{table*}

\begin{table*}[tp]
    \centering
    \caption{Pose results on our new dataset. Average accuracy (including error and jitter) will be evaluated only when its recall $> 10\%$. Best results are shown in bold and underlined.}
    \vspace{-0.1in}    
    \begin{tabular}{c|c|cc|cc|cc|cc|cc|cc}
    \hline \hline
    \multirow{3}{*}{Method} & \multirow{3}{*}{Marker} & \multicolumn{6}{c}{Normal ($\leqslant 50$cm and $\leqslant 60^{\circ}$) }&\multicolumn{6}{|c}{ Difficult ($\textgreater 50$cm or $\textgreater 60^{\circ}$)}\\ \cline{3-14}
    &&\multicolumn{2}{c}{Position (mm)}&\multicolumn{2}{|c}{Rotation (deg)}&\multicolumn{2}{|c}{Detection (\%)}&\multicolumn{2}{|c}{Position (mm)} &\multicolumn{2}{|c}{Rotation (deg)}&\multicolumn{2}{|c}{Detection (\%)}\\ \cline{3-14}
    &  & Error & Jitter & Error & Jitter & Recall & Prec. & Error & Jitter & Error & Jitter& Recall   & Prec.   \\ \hline
    \multirow{5}{*}{Original}  &TopoTag & 0.918 & 0.091 & 1.194 & 0.176 &	86.86 & 100 & - & - & - & - & 5.18 & 74.56\\ 
    &RuneTag$^{(+)}$ & - & - & - & - & 0 & 0 & - & - & - & - & 0 & 0\\ 
    &ARToolKitPlus & 6.876 & 0.185 & 4.788 & 0.371 & 100 & 100 & 17.664 & 0.913 & 21.700 & 3.618 & 51.95 & 100\\ 
    &ArUco  & 3.088 & 0.331 & 3.362 & 1.378 & 95.63 & 100 & 10.908 & 2.898 & 13.787 & 4.485 & 43.52 & 100\\ 
    &AprilTag   & 0.390 & 0.072 & 0.770 & 0.096 & 100 & 100 & 3.645 & 1.119 & 4.566 & 1.951 & 64.83 & 99.98\\ 
    \hline
    \multirow{7}{*}{DeepTag}  &TopoTag & 0.803 & \topone{0.029} & 0.355 & 0.161 & 100 & 100 & 0.889 & \topone{0.133} & \topone{0.228} & \topone{0.203} & 59.86 & 96.74\\ 
    &RuneTag$^{(+)}$ & 0.503 &  0.146 & 1.608 & 0.241 & 65.49 & 100 & - & - & - & - & 0.75 & 100 \\ 
    &ARToolKitPlus & 0.700 & 0.043 & 0.701 & 0.207 & 100 & 100 & 1.010 & 0.182 & 0.566 & 0.374 & 67.78 & 100 \\ 
    &ArUco & 0.231 & 0.047 & \topone{0.234} & \topone{0.044} & 100 & 100 & \topone{0.628} & 0.303 & 0.716 & 0.531 & 71.69 & 100 \\ 
    &AprilTag & 0.370 & 0.042 & 0.279 & 0.072 & 100 & 100 & 0.937 & 0.298& 0.462 & 0.230  & 76.14 & 100 \\ \cline{2-14} 
    &AprilTag - XO & \topone{0.161}& {0.041} & {0.253} & 0.099& 99.97 & 100 & 2.211 & 0.336 & 2.778  & 0.223 & 65.94 & 100\\ 
    &AprilTag - XA & 0.287 & 0.043 & 0.339 & 0.085 & 100 & 100 & 2.020 & 0.321 & 1.763 & {0.207} & 64.57 & 100\\ \hline
    Improvement & Best vs. Best & 59\% $\Uparrow$ & 60\% $\Uparrow$ & 70\% $\Uparrow$ & 54\% $\Uparrow$  & \multicolumn{2}{c|}{ \slashbox{}{}}  & 83\% $\Uparrow$  & 85\% $\Uparrow$ & 95\% $\Uparrow$ & 90\% $\Uparrow$ &  \multicolumn{2}{c}{ \slashbox{}{}}\\


    \hline\hline
    \end{tabular}

    \label{table:linseq_dataset}
\end{table*}

\begin{figure*}[tp]
	\centering
	\includegraphics[width=0.99\textwidth]{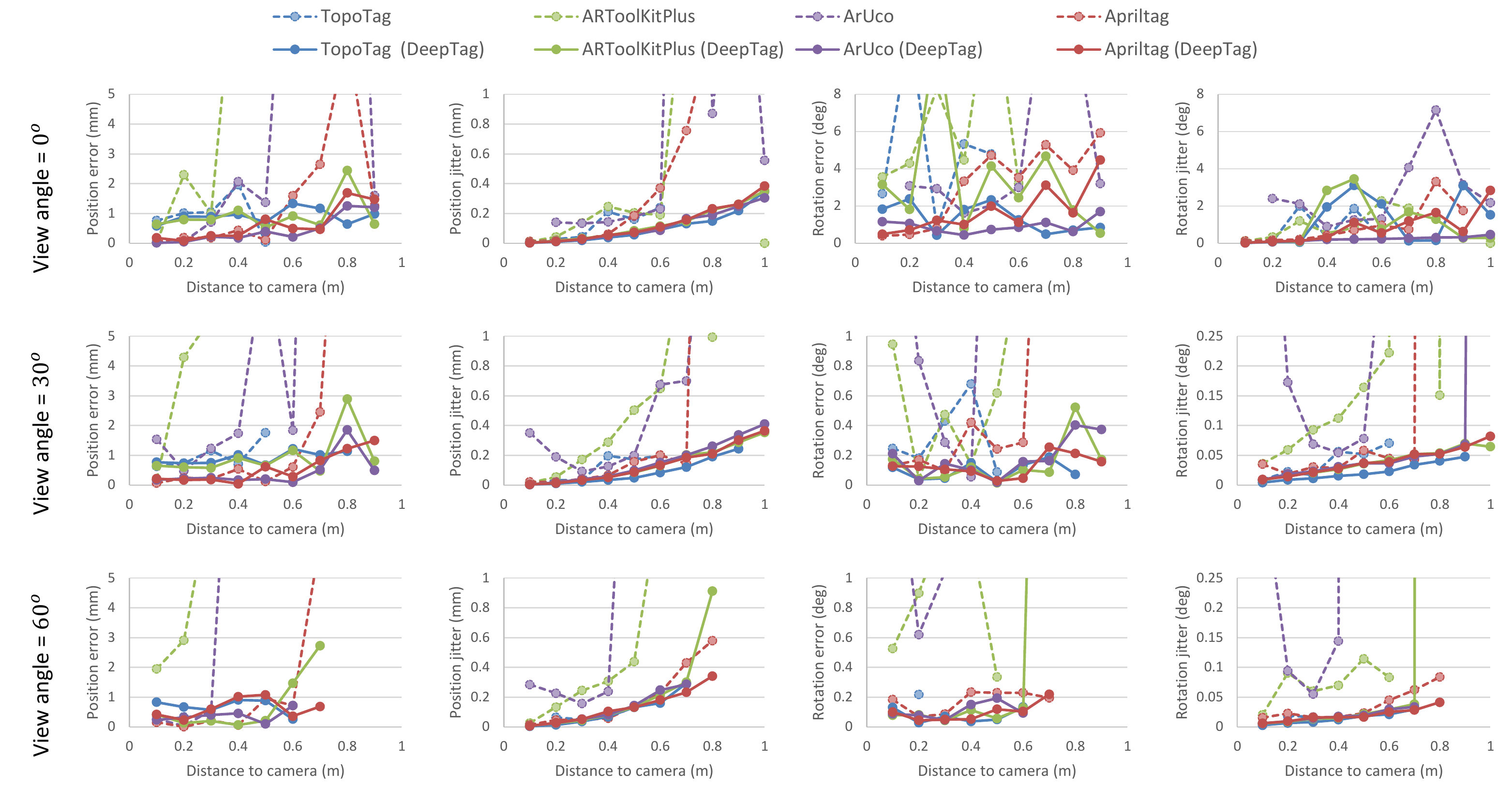}
	\caption{Pose accuracy results when markers are viewed at different distances and view angles. DeepTag consistently achieves equal or better results over original methods across all distances and view angles, and the improvement becomes larger as distance or view angle increases.}
	\label{fig:linseq_accuracy_plot}
\end{figure*}

Fig.~\ref{fig:linseq_accuracy_plot} further reveals how pose accuracy varies when markers are viewed at different distances and view angles.
DeepTag consistently achieves equal or better results over original methods across all distances and view angles, and the improvement becomes more significant as distance or view angle increases.
As view distance increases, as expected, pose accuracy starts to decrease for all methods. It's interesting to find that all methods achieve better rotation accuracy as view angle increases. This phenomenon is also found in previous work \cite{bergamasco2016an} due to better constraint of the reprojection.

\begin{figure*}[tp]
	\centering
	\includegraphics[width=0.99\textwidth]{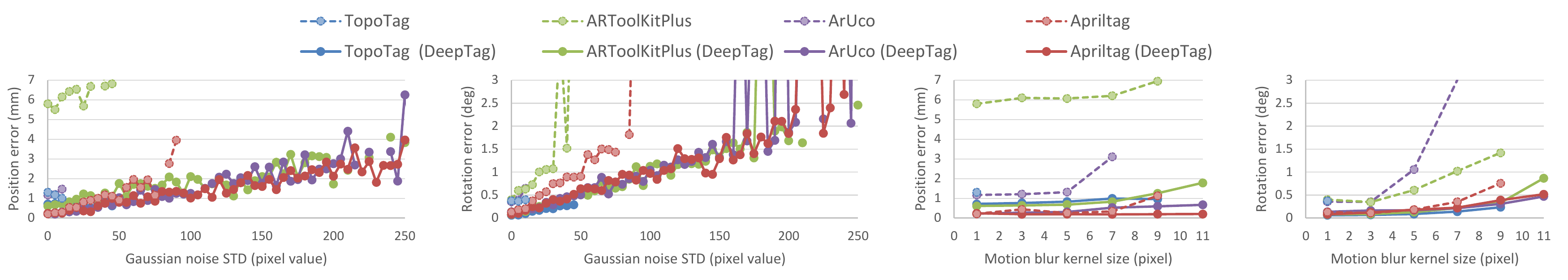}
	\caption{Pose accuracy results under Gaussian noise and motion blur. DeepTag greatly outperforms the original methods under the same level of Gaussian noise or motion blur.}
	\label{fig:linseq_accuracy_plot_gnoise_mblur}
\end{figure*}

\textbf{Noise and motion blur.}
Under a certain level of noise or motion blur, we estimate pose offsets for all the 100 possible combinations among the 20 images that are also used in previous noise and motion blur test, and average error of position and rotation will be evaluated.
As shown in Fig.~\ref{fig:linseq_accuracy_plot_gnoise_mblur}, as expected, pose error increases when Gaussian noise or motion blur becomes severe for all the methods.
Nevertheless, DeepTag greatly outperforms the original methods under the same level of Gaussian noise or motion blur.

\subsection{Further Discussion}

\textbf{3D marker support.}
Most existing markers are planar due to easy production, and from above we have demonstrated that DeepTag effectively supports planar markers with various designs.
On the other hand, as illustrated in Fig.~\ref{fig:linseq_accuracy_plot}, rotation error becomes more significant when the view angle is close to zero degree due to the lack of reprojection constraint.
To alleviate this issue, DeepTag can be extended to non-planar markers by adjusting ROI definition and the keypoint sorting scheme accordingly.
Another straightforward solution is to attach multiple planar to a 3D object like a cube or rhombicuboctahedron, see Fig.~\ref{fig:detect_cube} for examples.
In real applications, relative poses among these markers can be acquired via a calibration procedure.
A unified pose can then be estimated using all the keypoints of visible markers after detecting them separately.

\begin{figure*}[tp]
	\centering
	\includegraphics[width=0.99\textwidth]{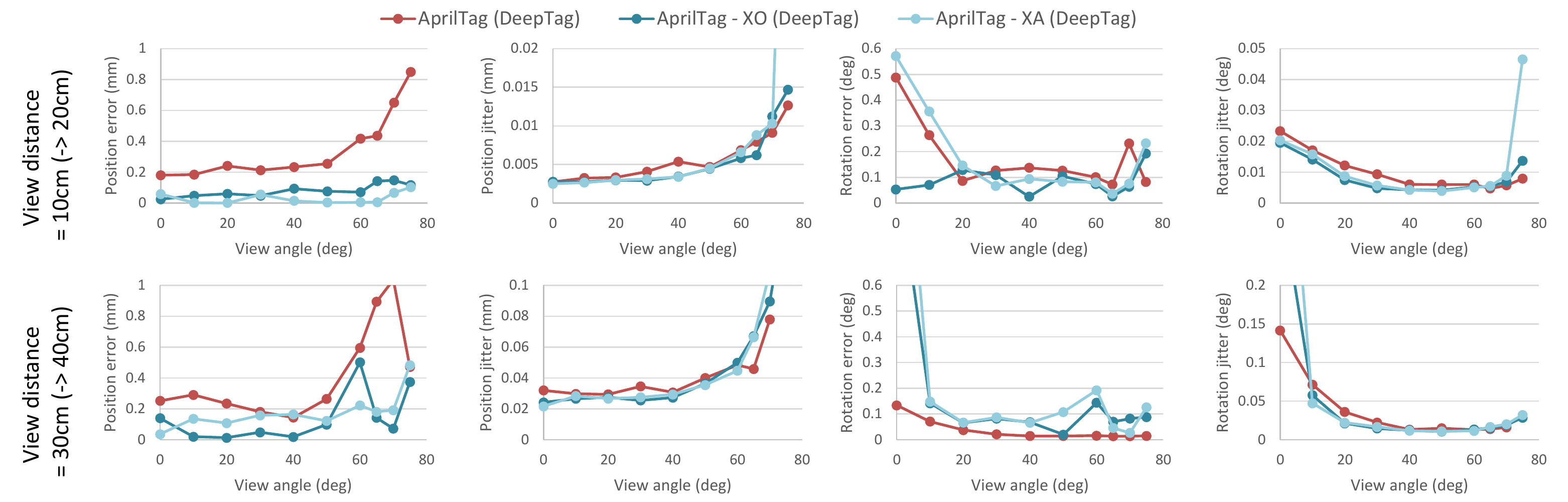}
	\caption{Validation results in two different view distances on the new dataset. The result of a location is plot if the recall is larger than 30\%. The table lists the results of AprilTag, AprilTag-XO, and AprilTag-XA achieved with DeepTag.}
	\label{fig:linseq_accuracy_plot_apriltag}
\end{figure*}

\textbf{Pattern selection.}
Fig.~\ref{fig:linseq_accuracy_plot_apriltag} is a detailed comparison of DeepTag performance for a given marker with three different patterns (i.e., original AprilTag marker and two our newly-designed AprilTag-XO, AprilTag-XA with complex patterns). Experiment is conducted at two different view distances (i.e., 10cm and 30cm), where example images are shown in Fig.~\ref{fig:linseq_image_examples}(a-b).
In a close view distance of 10cm, AprilTag-XO and AprilTag-XA achieve better pose accuracy over AprilTag for all four metrics, while rotation error becomes worse at 30cm though the conclusion of the other three metrics still holds.
This suggests that, for use scenarios where image quality is good enough (e.g., high resolution, markers usually in close range, etc.), markers with complex patterns are preferred for better pose accuracy.

Which is the best marker is still an open problem.
Fortunately, DeepTag supports design and detection of different marker families, and network training does not require additional data collection.
An optimal marker design for a certain scenario may need be determined by comparing the validation results of several possible marker designs.

\textbf{Timing and speedup.}
DeepTag is implemented using PyTorch~\cite{paszke2017automatic}, and tested on a PC with an Intel i7-11700K @ 3.60GHz CPU, 64GB RAM, and an NVIDIA GeForce GTX 3090 GPU.
CNN prediction can be accelerated with GPU while postprocessing on CPU.
In Stage-2, multiple ROIs are parallelly processed by CNN and serially postprocessed.
Computational times are recorded for input images of different resolutions.
For each resolution, we use the average time of 100 images.

Timing statistics are summarized in \autoref{table:timing_statistics}.
For an image with a single marker, DeepTag achieves 23fps with GPU for resolution of $640\times480$ and $256\times256$ in Stage-1 and Stage-2.
Comparing with the CNN prediction time and the total time, postprocessing is fast in Stage-1 thanks to the sparsity of the valid bounding boxes responded by CNN.
In contrast, postprocessing in Stage-2, including keypoint sorting and pose estimation for each valid marker, is much slower than CNN prediction.
Fortunately, postprocessing is not necessary for false positive ROIs by checking the total number of predicted keypoints.
Furthermore, with parallelization on GPU, CNN prediction for four ROIs is as fast as one single ROI.

\begin{table}[tp]
    \centering
    \caption{\REVISED{ Timing statistics of DeepTag (in sec.).}}
    \begin{tabular}{c|c|c|cc|cc}
        \hline \hline
                                    &\multirow{2}{*}{Resolution} & \multirow{2}{*}{\#} & \multicolumn{2}{c}{CPU}  & \multicolumn{2}{|c}{GPU}       \\ \cline{4-7}
                                    &                   &                       & CNN & Total & CNN & Total\\\hline
        \multirow{2}{*}{Stage-1}    &$640\times480$     & \multirow{2}{*}{1}    & 0.297 & 0.313 & 0.013 & 0.025 \\
                                    &$1280\times960$    &                       & 1.254 & 1.269 & 0.034 & 0.045 \\ \hline
        \multirow{4}{*}{Stage-2}    &\multirow{4}{*}{$256\times256$ }   & 1     & 0.064 & 0.078 & 0.007 & 0.019  \\ 
                                    &                                   & 2     & 0.117 & 0.146 & 0.007 & 0.031 \\ 
                                    &                                   & 4     & 0.211 & 0.274 & 0.007 & 0.057 \\ 
                                    &                                   & 8     & 0.414 & 0.555 & 0.011 & 0.108 \\ 
        \hline\hline
    \end{tabular}
    \label{table:timing_statistics}
\end{table}

When processing a continuous video sequence, tracking can be conducted with stage-1 or stage-2 alone.
As shown in Fig.~\ref{fig:linseq_accuracy_plot_ablation}, Stage-1 already achieves better pose accuracy than the original ARToolKitPlus algorithm.
Once the marker IDs are determined, it is feasible to track multiple markers in the subsequent frames using Stage-1 alone.
The computational time is independent of the number of markers, and 40fps can be achieved on $640\times480$ images with GPU.
On the other hand, to use Stage-2 alone, ROI input can be predicted from the marker poses in previous frames.
With an input resolution of $256\times256$, 53fps is achieved for a single marker.

\vspace{0.1in}

\begin{table}[tp]
    \centering
    \caption{\REVISED{Comparison of timing statistics (in sec.) under different noise levels.}}
    \REVISED{
    \begin{tabular}{c|ccccc}
        \hline \hline
        \multirow{2}{*}{Method}& \multicolumn{5}{c}{Gaussian Noise STD} \\ \cline{2-6}
                       & 0       &   10      & 30       & 100       & 300       \\ \hline
        TopoTag        & 0.0143   & 1.9603 & 8.9799    & 14.0633    & 11.2294   \\
        ARToolKitPlus  & 0.0017   & 0.0018  &  0.0024        & 0.0061 & 0.0097          \\
        ArUco          & 0.0043  &  0.0291  & 0.0378         & 0.0436   & 0.0415           \\
        AprilTag       & 0.0107  & 0.0181 &  0.0183  & 0.0101     &  0.0182    \\
        DeepTag        & 0.0649  & 0.0664     & 0.0674     & 0.0627     & 0.0646     \\
        \hline\hline
    \end{tabular}
    }
    \label{table:timing_statistics_compare}
\end{table}
\REVISED{
We also tested DeepTag on images with different levels of noise, and compare the computational time with that of the existing methods.
The test images contain a single marker and are with the resolution of $1280\times960$, and the timing statistics are shown in \autoref{table:timing_statistics_compare}.
The computational time of the existing methods generally increases when the noise level increases as the detection relys on the low-level elements instead of high-level ones. 
In contrast, as DeepTag depends on high-level detection, the computational time is not affected by noise. 
Note that, currently, DeepTag is implemented with Python and accelerated with GPU while the existing methods are implemented with C++ and tested on CPU. 
}



\begin{figure}[tp]
	\centering
	\includegraphics[width=0.5\textwidth]{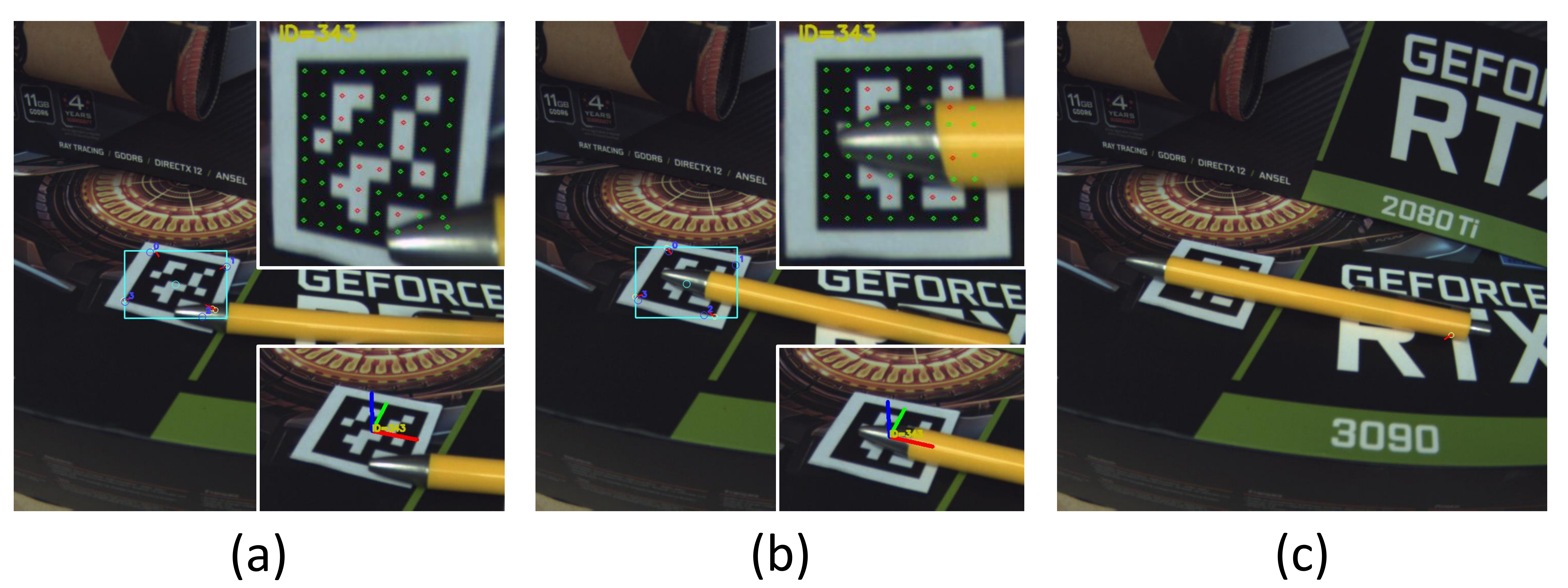}
	\caption{Detection results under occlusion. ROIs and corners of Stage-1 are drawn on the original images. Predicted keypoints and poses of Stage-2 are illustrated in the cropped patches. (a-b) DeepTag can tolerate small and moderate occlusion. (c) DeepTag fails under large occlusion.}
	\label{fig:limitation_occlusion}
\end{figure}

\textbf{Limitation.}
Occlusion is not specially taken into consideration in current DeepTag implementation.
As an example, Fig.~\ref{fig:limitation_occlusion} illustrates detection results of DeepTag under occlusion.
Thanks to the robustness of Stage-2 and the redundancy of keypoints, DeepTag can tolerate occlusion on a small part of the marker (Fig.~\ref{fig:limitation_occlusion}(a)), and still works well under moderate occlusion as long as the number of incorrect decoded digital symbols is within the Hamming distance allowed by the marker library  (Fig.~\ref{fig:limitation_occlusion}(b)).
When a large part of the marker is occluded, DeepTag would fail as Stage-1 fails to generate a response for the marker (Fig.~\ref{fig:limitation_occlusion}(c)).

It's worth noting that, the original AprilTag method fails to handle all these three cases shown in Fig.~\ref{fig:limitation_occlusion} as either the boundary or the inner bits cannot be successfully detected.
How to handle large occlusion is still an open problem for all fiducial marker systems, which requires the system with the ability to recover the whole structure from very few visible parts.

\vspace{0.1in}
\section{Conclusion}
We propose DeepTag, a general framework for fiducial marker design and detection.
DeepTag supports a large variety of existing marker families and achieves consistently better detection robustness and higher pose accuracy than traditional algorithms, which is extensively evaluated on both previous dataset and our new large challenging dataset.
Furthermore, DeepTag can easily adapt to new design and detection of new marker families with customized local patterns.
To facilitate broader applications, we further propose a synthesis procedure that generates training data on the fly without manual annotation. 

Which is the best marker design is still an open problem.
For future research, we will explore automatic marker design under different use scenarios.
With parameterized pattern generation procedure, the design and detection algorithms may be automatically trained and validated.

\appendices
%



\ifCLASSOPTIONcaptionsoff
  \newpage
\fi

\bibliographystyle{IEEEtran}
\bibliography{deeptag}




%



%

\begin{IEEEbiography}[{\includegraphics[width=1in,height=1.25in,clip,keepaspectratio]{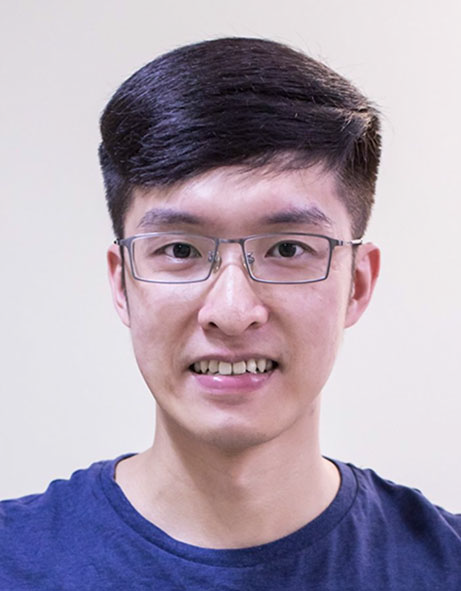}}]{Zhuming Zhang}
received the B.Eng. degree from Guangdong University of Technology, Guangzhou, China, in 2010, and the M.Eng. degree from South China University of Technology, Guangzhou, China, in 2013, and the M.Phil. degree from The Chinese University of Hong Kong, Hong Kong, China, in 2019.

He is currently with Guangdong Virtual Reality Co., Ltd. (aka. Ximmerse) as an algorithm engineer. His research interests include computer vision, machine learning, augmented reality and virtual reality.
\end{IEEEbiography}

\begin{IEEEbiography}[{\includegraphics[width=1in,height=1.25in,clip,keepaspectratio]{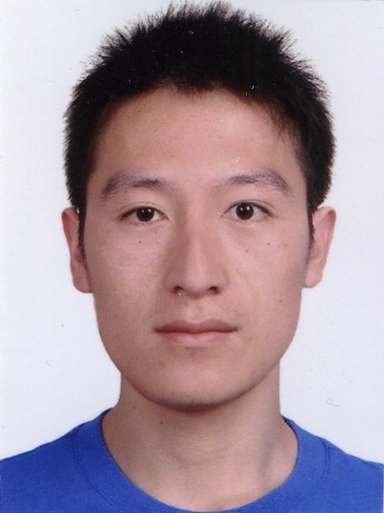}}]{Yongtao Hu}
received the B.Eng degree in computer science from Shandong University, Jinan, China, in 2010, and the Ph.D. degree in computer science from The University of Hong Kong, Hong Kong, in 2014.

He is currently with Guangdong Virtual Reality Co., Ltd. (aka. Ximmerse) as a research scientist. Prior to joining Ximmerse, he was a staff researcher with Image and Visual Computing Lab (IVCL), Lenovo Research, Hong Kong from Jan. 2015 to Oct. 2015, was a researcher assistant with IVCL from Jul. 2014 to Nov. 2014, and was a research intern at Internet Graphics Group in Microsoft Research Asia (MSRA) from Mar. 2010 to Jun. 2010. His research interests include computer vision, multimedia, machine learning, augmented reality and virtual reality.
\end{IEEEbiography}

\begin{IEEEbiography}[{\includegraphics[width=1in,height=1.25in,clip,keepaspectratio]{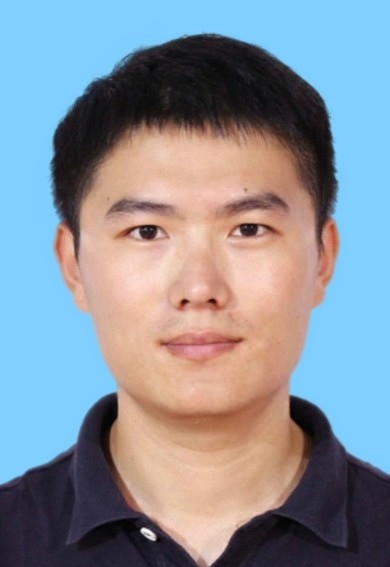}}]{Guoxing Yu}
received the B.Eng degree in electronic information engineering from Wuhan University of Science and Technology, Wuhan, China, in 2013, and the M.E. degree in information and communication engineering from Huazhong University of Science and Technology, Wuhan, China, in 2016.

He is currently with Guangdong Virtual Reality Co., Ltd. (aka. Ximmerse) as an algorithm engineer. Prior to joining Ximmerse, he was an algorithm engineer with Wuhan Guide Infrared Co., Ltd. Wuhan from Jul. 2016 to Aug. 2017. His research interests include computer vision, augmented reality and virtual reality.
\end{IEEEbiography}

\begin{IEEEbiography}[{\includegraphics[width=1in,height=1.25in,clip,keepaspectratio]{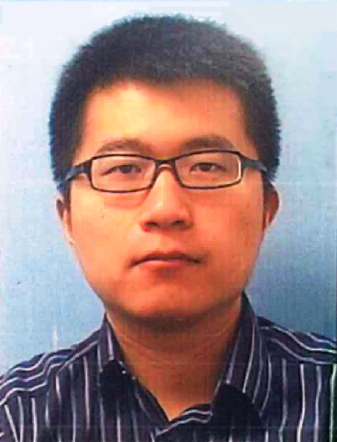}}]{Jingwen Dai (S'09 - M'12)}
received the B.E. degree in automation from Southeast University, Nanjing, China, in 2005, the M.E. degree in automation from Shanghai Jiao Tong University, Shanghai, China, in 2009, and the Ph.D. degree in mechanical and automation engineering from the Chinese University of Hong Kong, Hong Kong, in 2012.

He is currently with Guangdong Virtual Reality Co., Ltd. (aka. Ximmerse) as co-founder and chief technology officer. Prior to joining Ximmerse, he was a manager and advisory researcher with Image and Visual Computing Lab (IVCL), Lenovo Research, Hong Kong from Jan. 2014 to July 2015, and was a Post-Doctoral Research Associate with the Department of Computer Science, University of North Carolina at Chapel Hill, Chapel Hill, NC, USA from Oct. 2012 to Dec. 2013. His current research interests include computer vision and its applications in human-computer interaction, augmented reality and virtual reality.
\end{IEEEbiography}

%
%




\end{document}